\documentclass[5p,twocolumn]{elsarticle}

\usepackage{color,soul}
\usepackage[utf8]{inputenc}
\usepackage{xcolor}
\usepackage{listings}
\usepackage{multirow}
\usepackage{colortbl}
\usepackage{setspace}
\usepackage{comment}
\usepackage{url}
\usepackage{hyperref}
\usepackage{booktabs} 
\usepackage{tabu} 
\usepackage{tabularx}
\usepackage[T1]{fontenc}
\usepackage[scaled=0.9]{beramono}
\usepackage{microtype}
\usepackage[justification=centering]{caption}

\usepackage{enumitem}
\setitemize{noitemsep,,parsep=0pt,partopsep=0pt}
\journal{Journal of \LaTeX\ Templates}

\begin{document}

\begin{frontmatter}

\title{Enabling FAIR Research in Earth Science through Research Objects}
%\tnoteref{mytitlenote} (12-18 pages, Deadline May 30th)}
%\tnotetext[mytitlenote]{Obviously this is not the definitive title}


\author[expertsystemaddress]{Andres Garcia-Silva\corref{mycorrespondingauthor}} 
\cortext[mycorrespondingauthor]{Corresponding author}
\ead{agarcia@expertsystem.com}

\author[expertsystemaddress]{Jose Manuel Gomez-Perez\corref{mycorrespondingauthor}}
\ead{jmgomez@expertsystem.com}

\author[pscnaddress]{Raul Palma}
\ead{rpalma@man.poznan.pl}

\author[pscnaddress]{Marcin Krystek}
\ead{mkrystek@man.poznan.pl}

\author[meeoaddress]{Simone Mantovani}
\ead{mantovani@meeo.it}

\author[cnraddress]{Federica Foglini}
\ead{federica.foglini@bo.ismar.cnr.it}

\author[cnraddress]{Valentina Grande}
\ead{valentina.grande@bo.ismar.cnr.it}

\author[cnraddress]{Francesco De Leo}
\ead{francesco.deleo@bo.ismar.cnr.it}

\author[ingvaddress]{Stefano Salvi}
\ead{stefano.salvi@ingv.it}

\author[ingvaddress]{Elisa Trasati}
\ead{elisa.trasatti@ingv.it}

\author[ingvaddress]{Vito Romaniello}
\ead{vito.romaniello@ingv.it}

\author[esrinaddress]{Mirko Albani}
\ead{Mirko.Albani@esa.in}

\author[esrinaddress]{Cristiano Silvagni}
\ead{Cristiano.Silvagni@esa.int}

\author[esrinaddress]{Rosemarie Leone}
\ead{rosemarie.leone@esa.int}

\author[terradueaddress]{Fulvio Marelli}
\ead{fulviomarelli@me.com}

\author[satcenaddress]{Sergio Albani}
\ead{Sergio.Albani@satcen.europa.eu}

\author[satcenaddress]{Michele Lazzarini}
\ead{Michele.Lazzarini@satcen.europa.eu}

\author[bgsaddress]{Hazel J. Napier}
\ead{hjb@bgs.ac.uk}

\author[bgsaddress]{Helen M. Glaves}
\ead{hmg@bgs.ac.uk}

\author[bgsaddress]{Timothy Aldridge}
\ead{Timothy.Aldridge@hsl.gsi.gov.uk}

\author[unavcoaddress]{Charles Meertens}
\ead{chuckm@unavco.org}

\author[unavcoaddress]{Fran Boler}
\ead{boler@unavco.org}

\author[batelleaddress]{Henry W Loescher}
\ead{hloescher@battelleecology.org}

\author[batelleaddress]{Christine Laney}
\ead{claney@battelleecology.org}

\author[batelleaddress]{Melissa A Genazzio}
\ead{mgenazzio@battelleecology.org}

\author[sandiegoaddress]{Daniel Crawl}
\ead{crawl@sdsc.edu}

\author[sandiegoaddress]{Ilkay Altintas}
\ead{altintas@sdsc.edu}
 
\address[expertsystemaddress]{Expert System, Calle Profesor Waskman 10, 28036 Madrid}
\address[pscnaddress]{Poznań Supercomputing and Networking Center PSCN, Jana Pawła II 10, 61-139 Poznań, Poland}
\address[meeoaddress]{Meteorological and Environmental Earth Observation MEEO, Viale Volano 195/A Int. 2 I-44123 Ferrara, Italy}
\address[cnraddress]{Istituto di Scienze Marine-Consiglio Nazionale delle Ricerche ISMAR-CNR, Via Gobetti, 101 40129 Bologna Italia}
\address[ingvaddress]{Istituto Nazionale di Geofisica e Vulcanologia, Via di Vigna Murata, 605 00143 Roma Italy}
\address[esrinaddress]{European Space Agency ESA-ESRIN, Largo Galileo Galilei, 1, 00044 Frascati RM, Italy}
\address[terradueaddress]{Terradue Srl, Via Giovanni Amendola,46 00185 Rome Italy}
\address[satcenaddress]{European Union Satellite Center, Apdo. de Correos 511, 28850 Torrejón de Ardoz, 
Madrid – Spain}
\address[bgsaddress]{British Geological Survey, Nicker Hill, Keyworth, Nottingham NG12 5GG}
\address[unavcoaddress]{UNAVCO, Boulder, CO, USA}
\address[batelleaddress]{Battelle-National Ecological Observatory, Network, Boulder, CO, USA}
\address[sandiegoaddress]{San Diego Supercomputer Center, UCSD, La Jolla, CA, USA}
% * <jose.manuel.gp@gmail.com> 2018-05-24T09:58:43.535Z:
% 
% I don't like way email addresses look like as footnotes. Let's go back to showing them in the author list.
% 
% ^.


\begin{abstract}
%Scientific communities in data-intensive disciplines
Data-intensive science communities are progressively adopting FAIR practices that enhance the visibility of scientific breakthroughs and enable reuse. At the core of this movement, research objects contain and describe scientific information and resources in a way compliant with the FAIR principles and sustain the development of key infrastructure and tools. This paper provides an account of the challenges, experiences and solutions involved in the adoption of FAIR around research objects over several Earth Science disciplines.
%the creation of a FAIR research community, built around the concept of research objects, over several Earth Science disciplines. 
During this journey, our work has been comprehensive, with outcomes including: an extended research object model adapted to the needs of %the different communities of earth scientists; 
earth scientists; the provisioning of digital object identifiers (DOI) to enable persistent identification and to give due credit to authors; the generation of content-based, semantically rich, research object metadata through natural language processing, enhancing visibility and reuse through  recommendation systems and third-party search engines; and various types of checklists that provide a compact representation of research object quality as a key enabler of scientific reuse. All these results have been integrated in ROHub, a platform that provides research object management functionality to a wealth of applications and interfaces across different scientific communities. To monitor and quantify the community uptake of research objects, we have defined indicators and obtained measures via ROHub that are also discussed herein.
%last sentence i didnt change according to Hank, only clarified it a bit, as our goal is to quantify research object usage not ROHub usage, am i right?
\end{abstract}


\end{frontmatter}

%\linenumbers

\section{Introduction}
%Scientific communities in data-intensive 
Data-intensive science communities, together with a diverse group of stakeholders from academia, industry, funding agencies and publishers, are calling for innovative ways to manage their data, methods and other resources that enhance the visibility of scientific breakthroughs, encourage reuse, and foster a broader data accessibility\cite{bourne_et_al:DM:2012:3445}. These initiatives seek to overcome the current limitations imposed by conventional scholarly communications, %as well as the publication of data\footnote{Data Citation Synthesis Group Joint Declaration of Data Citation Principles: https://doi.org/10.25490/a97f-egyk} and research software\cite{10.7717/peerj-cs.86} in isolated repositories, which hinder scientific progress. 
such as the publication of data\footnote{Data Citation Synthesis Group Joint Declaration of Data Citation Principles: https://doi.org/10.25490/a97f-egyk} and research software\cite{10.7717/peerj-cs.86} in isolated repositories. 
%Hank comment is focusing only on data management
Modern science requires to systematically capture the research lifecycle %of scientific investigations 
%and to provide a unified entry point to information
and to provide a unified entry point with accepted (standardized) means to access the process-level information about %the hypotheses 
the scientific investigation, e.g., hypotheses investigated, the data used and produced %during experimentation or observation, 
in a study, the type of analytics and computations used, the derived conclusions, the researchers themselves, and the different versions and licensing of data or software, to name a few. Some envision a new science grand challenge: to create artificial intelligence that can eventually make major scientific discoveries worthy of a Nobel Prize \cite{DBLP:journals/aim/Kitano16}. %Still far from realization though, the latter highlights the increasing role of assisted means that support the scientific endeavor.
While this is still far from being realized, it highlights the increased need to enhance the type of data and knowledge management that supports the advancement of scientific frontiers\cite{Stocker:2017}.

%Research objects are one of the main enablers of such vision, with the potential to accelerate science and estimulate the uptake of good practices in data-intensive science. 
The use of Research Objects (RO) enable such vision, and have the potential to accelerate the production of scientific knowledge and foster the adoption of good data (and method) management practices. A research object \cite{Bechhofer2013599,BelhajjameEtAl:SePublica2012,conf/eScience/ZhaoGBKGGHRRG12} is a semantically enriched information unit that encapsulates all the materials and methods relevant to a scientific investigation, the associated annotations and the context where such resources were produced and used. Research objects can be viewed as technical and social artifacts with the goal to enhance the sharing, preservation and communication of data-intensive science, facilitate validation, and encourage reuse by the community. On the one hand, research objects address the technical challenges, e.g., preservation, reproducibility, and interoperability, and contain metadata that make them uniquely identifiable, processable, and machine readable. Inspired by software sustainability initiatives\cite{Crouch:2013:SSI:2719649.2719661,hettrick_2014_14809}, data, methods and software can be encapsulated as a citable research object in ways that are also complementary to traditional publications. On the other hand, research objects also address some of the social aspects %crucially involved 
in the scientific enterprise \cite{Barabasi639}, %facilitating that due credit is given to the authors of scientific contributions in their various forms, enabling discussion around the investigation, and ultimately supporting collaboration. 
by fostering author accreditation of their respective contributions that, in turn, enables personal and team advancement, discussion around the investigation itself, and supports collaboration and innovation. Moreover, there are other added benefits of the use of research objects in this context, such as broader distributions of the cited work, shortened publication times, and the release of other resources used in the scientific study. 

%As society and scholars move away from paper towards digital content, research objects have a key role to play in the way scientific results, methods and materials, are communicated, shared, and validated by the scientific community, given the need for mechanisms that support the production and reuse of self-contained, data-centric scientific products. Research objects encourage scientists to share in return of citations to their work, represented as a research object, also shortening publication times. In doing so, research objects support the creation of a virtuous circle of credit and attribution over the key computational resources involved in scientific research. Inspired in sustainable software development practices\cite{Crouch:2013:SSI:2719649.2719661,hettrick_2014_14809}, research objects encourage the release of scientific resources in addition to text publication, in the sense that data, methods and software can be encapsulated as a citable research object in complementary ways to traditional journal or conference publications.

Research objects reinforce %the vision contained in 
the FAIR Data Principles \cite{372751}: a concise and measureable set of guidelines %for those wishing to enhance the reusability of their data holdings. The FAIR Principles put specific emphasis on enhancing the ability of machines to automatically find and use the data, in addition to supporting its reuse by individuals. 
to enhance data reusability, which put emphasis on enhancing the ability of machines to automatically find and use data.
%Data being FAIR is also a way to support the '7-R's' 
Research objects also support the '7-R's' (Reusable, Repurposeable, Repeatable, Reproducible, Replayable, Referenceable, Respectful) that characterize reuse in e-laboratories \cite{Bechhofer2013599}, %initially motivated the creation of the research object concept. 
and which was the original motivation for the %creation of the research object concept.
creation of research objects in this context.
%The 7-R's fit into the FAIR principles and the desired scientific and research activities in which research objects play the key role. Indeed, as placeholders for everything releated to a scientific investigation, research objects provide a holistic approach towards the reuse of scientific knowledge: not only is data reusable but also put in the context of the investigation as a portable information artefact.
In support of the FAIR principles and 7-R's, research objects not only foster data reuse but also place the specific scientific study in a broader (accessible) perspective by providing contextual information about the study.

%This paper describes the journey of introducing research objects in Earth Science, from the understanding of the needs of these communities in terms of representing, disseminating and reusing scientific knowledge to the required extensions of the research object representation formalism and the associated infrastructure for research object management: ROHub\footnote{ROHub is available online at \url{http://www.rohub.org}}. The work described herein makes special emphasis on the exploitation of natural language processing and semantic annotation technologies to automatically generate research object metadata from their payload, producing richer, self-descriptive, expressive and machine-processable research objects while reducing human annotation effort, thus contributing to FAIR research and its reuse. In this paper we focus on the adoption of research objects by different scientific communities and disciplines in Earth Science, extending previous work in experimental sciences. The paper highlights the role of research objects in making research data FAIR and contextualized in the related scientific investigations and how this has a positive impact in the reuse of scientific knowledge and resources.

This paper describes the use of research objects in Earth Science, as an exemplar of the adoption of FAIR principles and the 7-R’s, supported by the ROHub platform \footnote{ROHub is available online at \url{http://www.rohub.org}}. Our approach has been informed and validated by numerous earth scientists from different communities, in the context of projects to build e-research infrastructure (EVER-EST project\footnote{\url{http://ever-est.eu/}}, and CoopEUS project\footnote{\url{https://www.neonscience.org/observatory/strategic-development/coopeus-project}}). We have set an ecosystem of tools around research objects that helps to ensure that they are rich in metadata, indexable, searchable and discoverable, authors receive due credit, and end-user applications are tailored to the needs of earth scientist.

The remainder of the paper is structured as follows. Section 2 further motivates our work by showing a FAIR assessment on available earth observation data sets. Section 3 describes research objects. Section 4 presents our approach to build a FAIR research environment. Section 5 describes the extensions, customizations, and enhancements to support end-user needs and the management of the research life cycle.  Section 6 focuses on the generation of content-based research object metadata. Section 7 shows how such metadata is leveraged by dedicated recommender systems and third party search engines.  In Section 8, we demonstrate our approach with 3 use cases. Section 9 presents our work towards community building and take up. Finally, Section 10 presents conclusions and future work.

%rap: I would change the title to something like: FAIR challenges in Earth Science, or merge in the intro.
\section{FAIR challenges in Earth Science}
Earth scientists work with heterogeneous datasets generated by data providers such as space agencies, specialized organizations and research projects that produce earth observation data. For example, scientists interested in marine litter need to 
% Hank comment change the meaning so i didnt include it...we assessed marine litter datasets to understand...
understand complex scientific inquiries about the 
distribution and sources of litter, the pathways, the transport mechanisms to the open deep sea, its transformations, the impact on the ecosystem and the sink of marine litter in the marine environment. For such tasks, they work with multiple data types, including: in situ sea floor observations from imaging technology (ROV or Dive transects), fishing trawling, geophysical surveys (e.g. Multi Beam and Side Scan Sonar), visual surveys of floating debris and data for oceanographic modeling. 

\begin{figure}[]
\centering
  \includegraphics[width=0.48\textwidth]{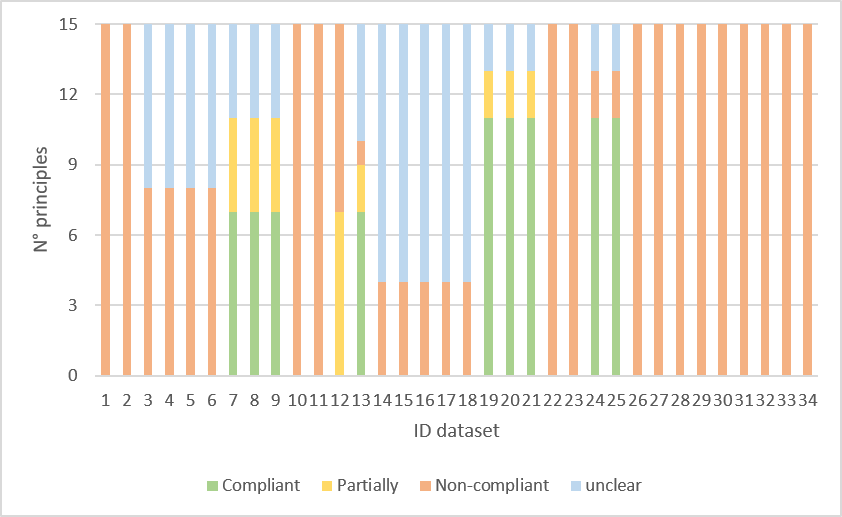}
  \caption{FAIR-ness evaluation of 35 datasets about marine litter}
  \label{fig:fairness}
\end{figure}

Were such data published according to the FAIR principles, it would be easier for domain scientists to focus exclusively on the analysis of the data and generate scientific results derived from such observations. However, this is typically not the case. We selected a sample of 35 highly curated, marine research datasets frequently used for marine litter analysis (table \ref{marinelitter} shows some of them), collected by public organizations and publicly funded research projects through EU framework programs and national programs and qualitatively assessed their level of FAIR-ness. To this purpose, we followed the methodology proposed by Dunning et al. \cite{dunning2017FairExcel}, which systematically evaluates each of the 15 principles corresponding to the 4 letters of FAIR. The methodology considers the information available on the website of the data provider, what is written on help pages, and what is visible in the published data record. The results of our analysis (see figure \ref{fig:fairness}) show that none of the selected datasets can be considered FAIR at the present stage, while most of them do not comply with the FAIR principles. While this analysis only covers a specific area of Earth Science, the conclusions we obtained illustrate the general situation of research data in the observational scientific disciplines.

% \usepackage{booktabs}
\begin{table}[]
\centering
\scriptsize
\tabcolsep=0.11cm
\caption{Shortlist of public marine litter data sets per project}
\label{marinelitter}
\begin{tabular}{@{}llllll@{}}
\toprule
\textbf{Project} & \textbf{Format} & \textbf{Size} & \textbf{Period} & \textbf{Area}                                                      \\ \midrule
HERMIONE         & .shp .csv       & 200 KB        & 2009-2012       & \begin{tabular}[c]{@{}l@{}}Artic, Atlantic, \\ Mediterranean\end{tabular} \\
PERSEUS          & .shp .csv       & 200 KB        & 2012-2015       & Mediterranean                                                                \\
MIDAS            & .shp .csv       & 200 KB        & 2013-2016       & Mediterranean                                                                \\
PROMETEO         & .mp4            & 1 GB          & 2007-2010       & Mediterranean                                                                \\
OASIS DEL MAR    & .mp4            & 1 GB          & 2010-2012       & Mediterranean                                                               	\\
Ritmare    		& .shp            & 14 MB          & 2013       & Venice Lagoon                                       								 \\
CoCoNet    		& .shp            & 90 GB          & 2012-2015  & Adriatic                                           			                    \\ \bottomrule
\end{tabular}
\end{table}

\section{Research Objects}\label{ro}

Research objects describe aggregations of scientific knowledge in a form, rich with annotations, that makes it recognizable, processable, and exchangeable by both humans and machines. A research object is a semantically rich aggregation of resources that bundles together essential scientific information about a scientific investigation \cite{Bechhofer2013599}. This information is not limited merely to the data used and the methods employed to produce and analyze such data, but it may also include links to the members of the investigation as well as other important metadata that describe the characteristics, inter-dependencies, context and dynamics of the aggregated resources \cite{Bechhofer2013599} \cite{BelhajjameEtAl:SePublica2012}. As such, a research object can encapsulate scientific knowledge and provide a mechanism for sharing and discovering reusable assets of the investigation within and across relevant communities, and in a way that supports the reliability and reproducibility of the results of such investigation. Nowadays, ROHub \cite{palma2014rohub} is the reference platform for research object management, with myExperiment as its nearest precursor \cite{Goble:2007:MSN:1273360.1273361}.

While there are no pre-defined constraints related to the type of resources that a research object can contain, in the context of scientific research the following usually apply:

\begin{itemize}
 \item Data used and produced during the experiment or observation.
 \item Scientific methods applied.
 \item Software and workflows implementing the methods.
 \item Provenance and execution settings.
 \item People involved in the investigation.
 \item Annotations about these resources, to interpret the scientific outcomes captured by a research object.
\end{itemize}   

\begin{figure*}[]
\centering
  \includegraphics[width=0.625\textwidth]{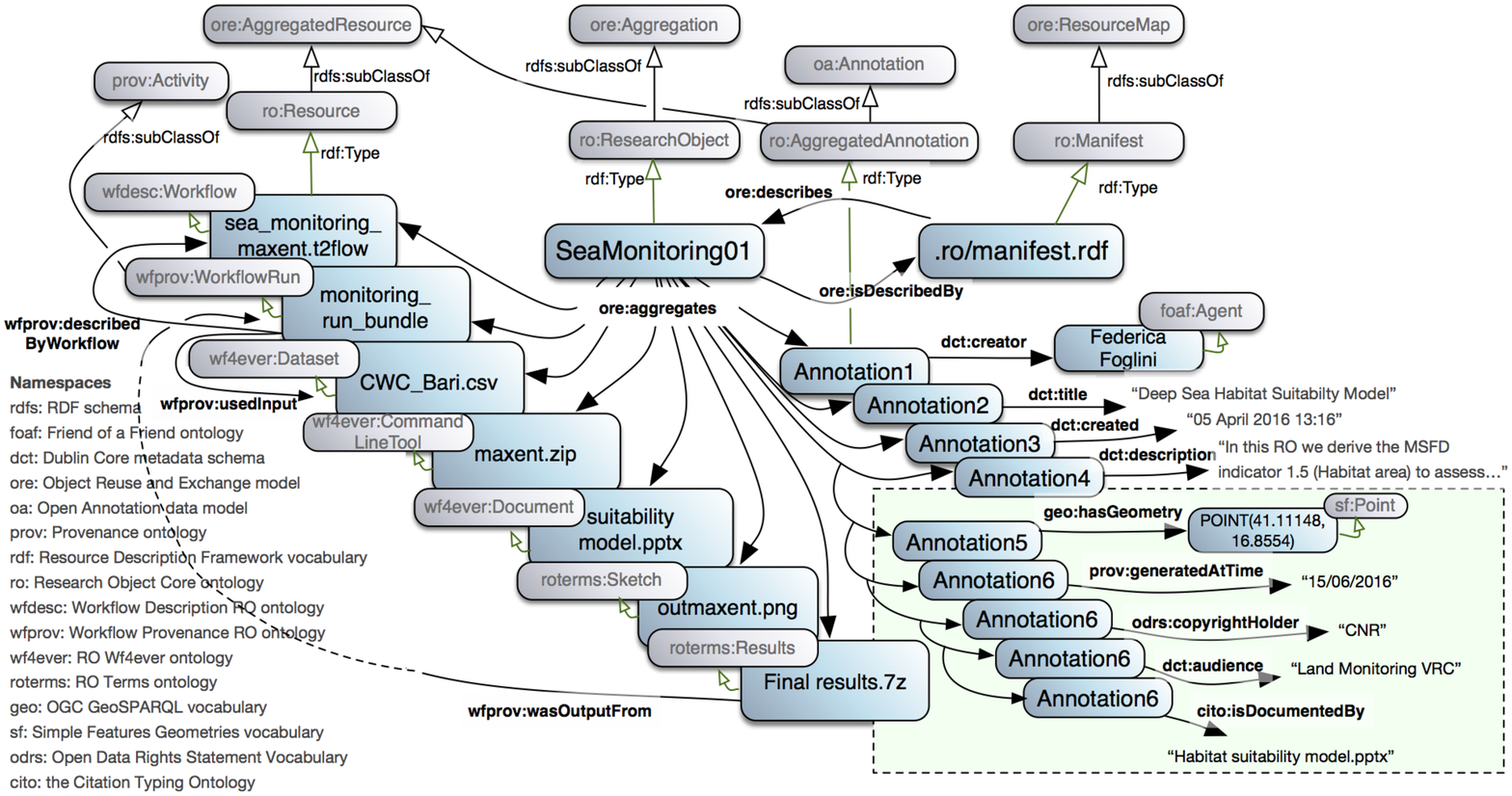}
  \caption{Simplified view of the research object containing a habitat suitability model (earth science specific metadata in the dashed rectangle).}
  \label{fig:ro}
\end{figure*}

The research object model relies on the W3C Resource Description Framework RDF \cite{w3c_rdf}, a data model specifically designed for data interchange in the web, and the Web Ontology Language OWL \cite{W3C04owl}, a rich knowledge representation model. In practice, this means that research objects can be easily processed not only by humans but also by machines, since both data and its semantics are described following standard means. The research object model comprises a set of vocabularies that allow describing a research object formally. Such vocabularies are defined in the following ontologies:

\begin{itemize}
 \item \textbf{The Research Object Core Ontology}\footnote{\url{http://purl.org/wf4ever/ro}} (ro), describing the aggregation of resources in the research object, as well as the annotations made on those resources. 
 \item \textbf{The Workflow Description Ontology}\footnote{\url{http://purl.org/wf4ever/wfdesc}} (wfdesc), meant as an upper ontology for more specific workflow definitions, and as a way to express abstract workflows.
 \item \textbf{The Workflow Execution Provenance Ontology}\footnote{\url{http://purl.org/wf4ever/wfprov}} (wfprov), for the representation of provenance information generated by the execution of a scientific workflow.
 \item \textbf{The Research Object Evolution Ontology}\footnote{\url{http://purl.org/wf4ever/roevo}} (roevo), which describes research object lifecycle information.
\end{itemize}   

Aggregation is supported through the use of the OAI-ORE vocabulary while annotation is supported by the Web Annotation Ontology\footnote{\url{Respectively, http://openarchives.org/ore} and \url{https://www.w3.org/ns/oa}}. In addition, the research object model makes use of existing vocabularies, in particular, Friend of a Friend (FOAF), Dublin Core Terms (DCTerms), and the Citation Typing Ontology (CITO), to provide research object authors with the means to express aspects such as the contributors to a research object, its citations, and the dependencies the research object and its content may have.

Figure \ref{fig:ro} shows a graphical representation of an existing research object \footnote{\url{http://sandbox.rohub.org/rodl/ROs/SeaMonitoring01/}} that uses the core vocabulary. This research object shows a partial and simplified view of the structure of an existing exemplary research object, which uses several modules of the research object ontology suite. It contains a habitat suitability model to derive the Marine Strategy Framework Directive indicator 1.5 (habitat area), assessing a descriptor of biological diversity. The research object encapsulates a scientific workflow, the input dataset, provenance information about the execution of the workflow, the output dataset, ancillary documentation such as images and presentations, and information regarding the author, plus metadata about the research object evolution and quality checks.

%\subsection{Vocabulary for Earth Science}\label{ro_es}
%Originally validated in experimental sciences \cite{Belhajjame201516}, in this paper we expand on this and report our experience bringing the research object concept to observational disciplines, particularly Earth Sciences. 

% rap: Hank proposes title RO Model functional Description, I think is not fitting well, but the current one is also too abstract. Maybe something like: "FAIR Ecosystem based on Research Objects"
%\section{Suite of models and tools for FAIR research}
\section{FAIR research environment based on Research Objects}
\label{sec:fair-research}
To enable a FAIR research environment we %propose an ecosystem of models and tools that interplay in order
advocate for the creation of an ecosystem of tools around the research object model and lifecycle. %added corresponding instead of used in sequence proposed by Hank (see coment below)
%to help scientists to easily share, find and reuse scientific results. These models and tools make up the ecosystem described here, which is depicted in figure \ref{fig:fairro}, for enabling data-intensive research communities like Earth Sciences to become FAIR.  
%The ecosystem depicted in figure \ref{fig:fairro} contextualizes the contributions presented in this paper in order to enable data-intensive research communities like Earth Sciences to become FAIR. 
% The figure 2 is generally useful, however, after seeing the comment from Hank, I think there is potential problem that these ecosystem components are seen as sequential or as a the study lifecycle, while I think they are not necesarly ordered. For me there is more like a loose dependency (in most cases) of one circle to the previous one. I am thinking of some other way of representing this. I see two possibilities: like a block diagram with layers (similar to architecture diagram), or as concentric circles. What do you think ?
The research object model is at the core of this ecosystem since it is based on a standardized formal semantics and an agreed upon vocabulary, making scientific outcomes interoperable, machine-readable, and shareable.
%We argue that the research object model is at the foundations of such transformation, since it defines an agreed vocabulary to share scientific outcomes that makes them interoperable and machine-readable, thanks to the use of a standard data format and formal semantics. 
The research object model is generic enough to accommodate any scientific community. Nevertheless, 
to make it practical for earth sciences we 
extended and customized the model to the specific needs of this area of science. %We interviewed and trained earth scientists from research organizations to identify such needs and extended the research object model accordingly. 

%HANK COMMENT ON FIGURE: The caption needs a tad more description, such as ‘ a conceptual duiagram that outlines the suite of RO models used in tandem, the flow of infromation, and therespective fucntions of each model.’
%RAP: I WOULD 
\begin{figure*}[htb]
\centering
  \includegraphics[width=0.60\textwidth]{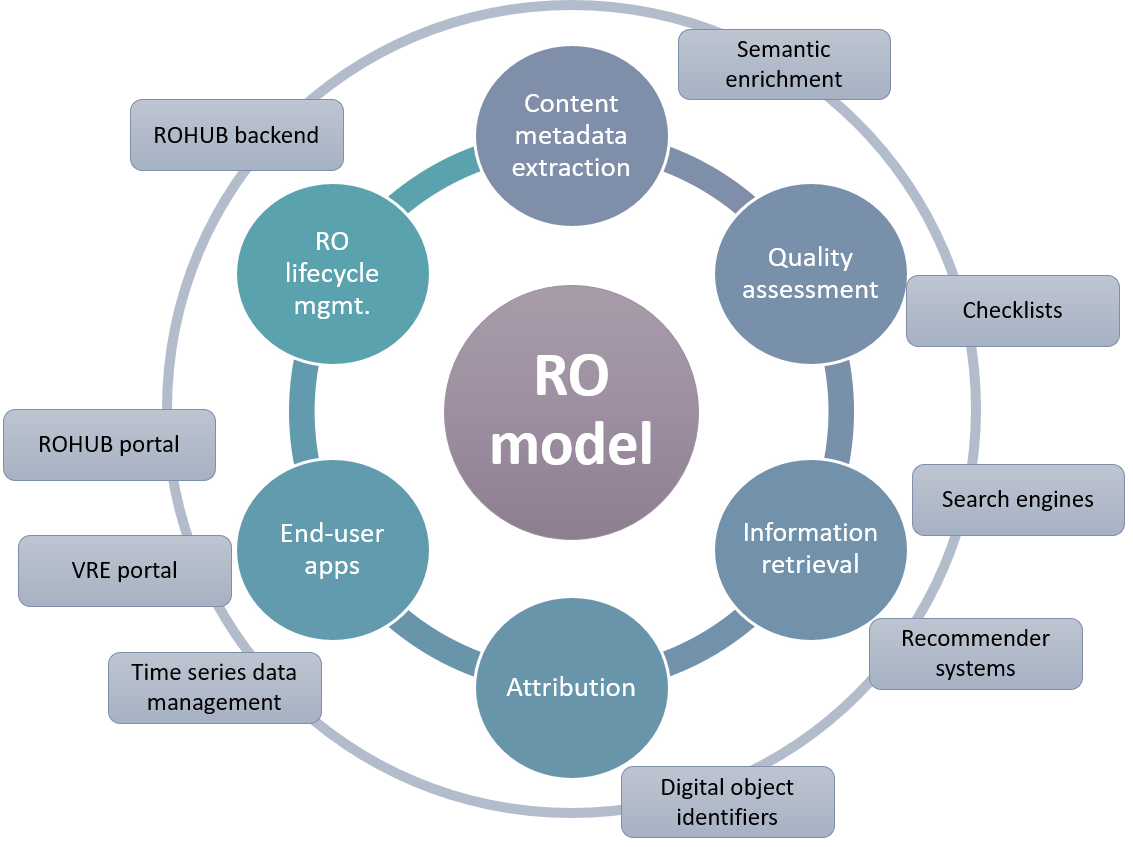}
% * <jose.manuel.gp@gmail.com> 2018-06-16T20:26:02.476Z:
% 
% This reminds me of the collaboration spheres :-) It conveys the idea clearly, though.
% 
% ^.
  \caption{Conceptual diagram that outlines the suite of tools around research objects to enable FAIR research. Around the research model the inner circle depicts the features required and the outer circle shows their technical support.}
  \label{fig:fairro}
\end{figure*}

The ecosystem, which is depicted in figure \ref{fig:fairro}, takes into account that: i) rich and expressive metadata is a key factor for sharing and reuse, ii) scientific results need to be visible and easily discovered, iii) scientists need to receive due credit for their work, and iv) research object management capabilities need to be integrated in existing analytic tools already in use by earth scientists in order to 
foster adoption. 

%scientific  applications need to embed analytic tools used by earth scientists and research object management to foster adoption.
% * <jose.manuel.gp@gmail.com> 2018-06-16T20:12:24.355Z:
% 
% > end-user applications have to embed analytic tools used by earth scientist and research object management to foster adoption.
% Not clear what this means. Don't like the way it is expressed eithers.
% 
% ^.

First, the research object model %enables users to produce metadata about the research object structure, content, and lifecycle. 
enables to capture specific metadata from each of the processes and tools used in the research lifecycle and ensembles them into a more comprehensive suite of metadata about the structure, content, and lifecycle of the research object. The structure and lifecycle metadata can be generated automatically by a research object management system, e.g., ROHub. %, assisting the task and relieving scientists from the burden of producing such metadata themselves. 
However, producing metadata about the content of a research object, e.g. unstructured text like scientific papers, slides, etc. is a complex tasks that requires more intelligent management of the information, which typically falls on the scientist.
%didnt change with Hank comment "system" as also changes meaning
%As a consequence, such key aspect of the metadata related to a scientific investigation is usually neglected and scarce. To generate content metadata we propose a semantic enrichment process that uses natural language processing against the research object payload to generate semantic metadata describing its content. 
As a consequence, these metadata are usually neglected and scarce. Our solution to this issue is a semantic enrichment process that carries out natural language processing against the research object payload. In addition, it is necessary to establish functionality that monitors the availability of current and relevant metadata, and the overall quality of the research object. We address this challenge through the use of checklists, defined according to the research object usage scenarios with the input of earth scientists.

Second, we make sure that research objects are indexable and searchable by search engines and tools that leverage the available metadata. Moreover, we developed a recommender system that identifies research objects that may be similar (in terms of their content) to other objects selected by a scientist. 

Third, dynamic accreditation is crafted through an extension of the research object lifecycle with a fork mechanism inspired by  software development practices \cite{10.1007/978-3-642-33442-9_1}, which automatically cites the research object being reused. Moreover, ROHub (a DataCite\footnote{\url{https://www.datacite.org}} member) assigns Digital Object Identifiers (DOI) to research objects upon release of intermediate or final research results. 
% * <jose.manuel.gp@gmail.com> 2018-06-16T20:23:00.984Z:
% 
% > dynamic
% What does this mean??
% 
% ^.

%On top of the model and the metadata, domain-specific applications allow researchers to easily produce and reuse research objects. Here, a core challenge is to integrate research objects with the tools and datasets that earth scientist use in their daily work. At the backbone of such  applications, ROHub supports CRUD operations and implements the research object lifecycle. On top of its back-end services and APIs, ROHub offers a generic research object management portal where scientists can create research objects and reuse existing ones from the repository, observing their access policies and licensing schemes. Additionally, earth scientists need specialized user interfaces adapted to earth observation work practices and specific types of relevant data, involving e.g. images, time series, and geolocalized data. In addition to the ROHub portal, in the paper we describe two additional user interfaces working on ROHub's back-end in different communities of earth scientists: a Virtual Research Environment that brings together earth observation datasets and processing tools to do research in Earth Sciences and share outcomes as reusable research objects, and a time series data management application where earth scientist can easily query and visualize on a map real-time data coming from data providers as UNAVCO (GPS data) and NEON (wind and humidity, among others), that can be sliced and stored along with provenance information in a research object. 

Lastly, end-users have a plethora of analytical tools tailored to their scientific disciplines.  Hence, the challenge is to develop functional capabilities to integrate research object with the tools and datasets that facilitate earth-system science, e.g., statistical packages, images, time series, remote sensed- mapped resources, and geo-referenced data, etc.  As a solution, ROHub offers a generic research object management portal where scientists can create research objects and reuse existing ones from its repository, and can manage their access policies, resources and metadata including licensing.  In addition, we describe two additional %functions
applications working on ROHub's back-end to facilitate the integration of other user interfaces: a Virtual Research Environment (VRE) that brings together earth observation datasets and analytical tools, and a time series data management application to more easily query and visualize real-time data on a map.

In table \ref{tab:faircomponents} we show how the contributions presented in this paper support FAIR research data in our target scientific communities. The research object model covers practically all the aspects of FAIR. Nevertheless, while the model enables the generation of FAIR data, tools that implement the model are required to actually  produce and manage FAIR data. DOIs, as permanent identifiers, reinforce findability and reuse given that they link to metadata about the publication. The semantic enrichment enhances findability by producing rich metadata, while checklists support accessibility, by validating that metadata is available, and reuse, by checking the existence of license and provenance metadata among other types. The visibility of research objects by search and recommendation systems is another step towards increasing findability. Finally, ROHub, which has been built on top of the the research object model and integrates the other developments presented herein, supports the generation and reuse of FAIR data. This is further illustrated by the other applications described in the paper, i.e. the EVER-EST virtual research environment and the time-series data management application developed in the context of the COOPEUS project.
\begin{table*}[t]
\centering
\scriptsize
\tabcolsep=0.11cm
\caption{Research object model and tools in support of FAIR principles. Rows are a subset of 12 FAIR principles and the model or tool support of each principle is indicated with an x.}
\label{tab:faircomponents}
\resizebox{\textwidth}{!}{
\begin{tabular}{|ll|c|c|c|c|c|c|c|c}
\hline
  \multicolumn{2}{|c|}{\begin{tabular}[c]{@{}c@{}}Models \& tools \textbackslash \\ Principles\end{tabular}} 
& \multicolumn{1}{|c|}{\begin{tabular}[c]{@{}c@{}}Research object model + \\ Earth Science \\ Extensions\end{tabular}} 
& \multicolumn{1}{c|}{\begin{tabular}[c]{@{}c@{}}Digital Object \\ Identifiers \\ (DOI)\end{tabular}} 
& \multicolumn{1}{c|}{\begin{tabular}[c]{@{}c@{}}Semantic Enrichment  \\ \& \\ Quality Assessment\end{tabular}} 
& \multicolumn{1}{c|}{\begin{tabular}[c]{@{}c@{}}Search Engines\\ \& \\ Recommenders\end{tabular}} 
& \multicolumn{1}{c|}{\begin{tabular}[c]{@{}c@{}}User interfaces: \\ ROHub portal\end{tabular}}  
& \multicolumn{1}{c|}{\begin{tabular}[c]{@{}c@{}}User interfaces: \\ VRE portal\end{tabular}} 
& \multicolumn{1}{c|}{\begin{tabular}[c]{@{}c@{}}User interfaces: \\ Time Series \\ Data Management\end{tabular}} 
\\ \hline                    
                    & - Rich Metadata                                                               & x                                                                                      &                                                                                                    &x                                                                                     &                                                                                                &x      &x                                                                      &x                                 \\
                    & - (meta)Data searchable                                                       &x                                                                                      &                                                                                                    &                                                                                                              &x                                                                       &x      &x                                                                      &x                               \\
\multirow{-3}{*}{F} & - Persistent Identifier                                                       &x                                                                                      &x                                                                           &                                                                                                              &                                                                                                &x      &x                                                                      &x 
\\ \hline
                    & - (meta)Data retrievable                                                      &x                                                                                      &x                                                                           &x                                                                                                             &                                                                                                &x      &x                                                                      &x                         \\
                    & - Open \& universal protocol                                                  &x                                                                                      &                                                                                                    &                                                                                                              &                                                                                                &x      &x                                                                      &x                               \\
\multirow{-3}{*}{A} & \begin{tabular}[c]{@{}l@{}}- Authentication \&\\   Authorization\end{tabular} &x                                                                                      &                                                                                                    &                                                                                                              &                                                                                                &x      &x                                                                      &x  
\\ \hline
                    & - Formal Knowledge Rep                                                        &x                                                                                      &                                                                                                    &                                                                                                              &                                                                                                &x      &x                                                                      &x                               \\
                    & - FAIR Vocabularies                                                           &x                                                                                      &                                                                                                    &                                                                                                              &                                                                                                &x      &x                                                                      &x                               \\
\multirow{-3}{*}{I} & - Link to other metadata                                                      &x                                                                                      &                                                                                                    &                                                                                                              &                                                                                                &x      &x                                                                      &x                               \\ \hline
                    & - Usage license                                                               &x                                                                                      &                                                                                                    &x                                                                                     &                                                                                                &x      &x                                                                      &x                                 \\
                    & - Provenance                                                                  &x                                                                                      &                                                                                                    &x                                                                                     &                                                                                                &x      &x                                                                      &x                                 \\
\multirow{-3}{*}{R} & \begin{tabular}[c]{@{}l@{}}- Standard community\\   meta(data)\end{tabular}   &x                                                                                      &x                                                                           &           x                                                                                                        &                                                                                                &x      &x                                                                      &x \\
\hline
\end{tabular}
}
\end{table*}

\section{Research Object Model - Earth Science Extensions}\label{ro-es}
In this paper we focus on scientific communities in Earth Science disciplines including sea monitoring, volcanology and biodiversity, that use earth observation data for different purposes. Such communities are represented by the following institutions.
\begin{itemize}
	\item \textbf{Institute of Marine Science} (CNR-ISMAR)\footnote{\url{http://www.ismar.cnr.it}}.
   	\item \textbf{Geohazard Supersites and Natural Laboratories} (GSNL)\footnote{\url{http://supersites.earthobservations.org}}, represented by the Italian National Institute of Geophysics and Volcanology (INGV).
\item \textbf{National Ecological Observatory Network} (NEON)\footnote{\url{https://www.neonscience.org/}}.
\end{itemize}

All these communities pursue FAIR practices for collaboration, sharing and reuse of scientific knowledge, even before actual publication of their work in conferences or journals. Two additional organizations focused on earth observation took part in our study, equally contributing requirements for the extension of the research object model and producing exemplary research objects: The UK Natural Hazards Partnership (NHP)\footnote{\url{http://www.naturalhazardspartnership.org.uk}}, and the European Union Satellite Centre (SatCen)\footnote{\url{https://www.satcen.europa.eu}}. However, while the former three are focused on scientific research missions (and therefore fall in the scope of this paper), the last two serve operational purposes, providing earth observation services to a limited set of stakeholders and security agencies. 

The research object model was developed initially in the context of experimental disciplines like genomics and astrophysics \cite{Belhajjame201516}, where scientific workflows play a central role to enable reproducibility. However, though that is also a relevant aspect for Earth Science communities, these are more focused on observations, e.g. involving the analysis of time series satellite data, rather than experimentation. Therefore we carried out a gap analysis to identify the necessary updates to be implemented in the model. In doing so, we used three main channels \cite{EVEREST:D4.1}:

\begin{itemize}
	 \item \textbf{A requirements questionnaire }with 14 questions related to the intended use of research objects that was distributed to each of the four organizations.
	\item \textbf{A survey }addressed to the broader Earth Science community containing a subset of the above questionary,  distributed ammong the participants of the Research Data Alliance RDA 9th Plenary Meeting\footnote{\url{https://www.rd-alliance.org/plenaries/rda-ninth-plenary-meeting-barcelona}}.
	\item \textbf{Two Research Object Hackathons}, where 50+ users in total from the four organizations received training on research objects methods and tools and started modeling their own exemplars. In the first hackathon, delegates from other scientific domains like Astrophysics\footnote{\url{http://www.iaa.es}} also participated, sharing their experiences with research objects.
\end{itemize}

The analysis of the surveys and the hackathons revealed five main areas where the gap between the coverage provided by the research object model and the needs of earth scientists were significant: geospatial information, time-period coverage, intellectual property rights, data access policies, and general-purpose information. In some cases, such information was not covered at all by the previous version of the research object model (geographic, time, data access policies), and in other cases it was not covered with sufficient detail as required by the earth scientists (intellectual property rights). The main additions to the model are summarized below (details available in this technical report \cite{EVEREST:D4.2}) and illustrated in Figure \ref{fig:ro} (see the annotations, and prefixes indicating the vocabularies used to model the new information, enclosed in the lower-right dashed rectangle). 

\begin{itemize}
	\item \textbf{Geospatial}, the coordinates of the region relevant for the research object and the observation it represents.
	\item \textbf{Time-period:} time span covered in the observation.
	\item \textbf{Intellectual property rights}, including copyright holder, copyright starting year, type of license and attribution.
	\item \textbf{Data access policy}, i.e. the access level and policies under which the research object can be accessed.
	\item \textbf{General metadata}, including the main scientific discipline of the research object, the size and format of the resources aggregated by the research object, the date when the research object was released, its digital object identifier (DOI), the status according to the research object lifecycle, and its target community.
\end{itemize}

The executable resources covered by the model have also been extended to cover not only scientific workflows but also other types of processes, such as web services, scripts,  command line tools and dedicated software frequently used in Earth Sciences. Earth scientist also requested new types of research objects according to the kind of the aggregated resources. We extended the research object types to characterize not only workflow-centric research objects, but also data-centric and service-centric, as well as documentation and bibliographic research objects. Finally, the research object lifecycle was extended with a new status (forked), which characterizes a new branch of the research object derived from the main one. 

While some of these changes were considered important for the overall research object community and were incorporated in the research object model\footnote{\url{https://github.com/ResearchObject/specifications/issues/13}}, other updates were specific to Earth Sciences. Therefore we created a new branch in the code repository of the research object model containing all the new metadata elicited in our analysis\footnote{\url{https://github.com/wf4ever/ro/tree/earth-science}}. %Again, we refer the reader to Everest deliverable 4.2 \cite{EVEREST:D4.2} for more details.

\subsection{Lifecycle Management Extensions}
\label{sec:lifecycle}

The lifecycle refers to the different stages that a scientific research (and its associated research object) transitions, from hypothesis generation to publication and archival. In previous versions of the research object lifecycle \cite{BelhajjameEtAl:SePublica2012}, research objects could be \textit{Live} (mutable research objects related to on-going research processes), \textit{Snapshot} (immutable research objects derived from live research objects, that are ready to release intermediate results), and \textit{Archived} (immutable research objects with final research results, where the research process has been completed). However, the creation of snapshots and archived research objects was limited to the authors of the particular research object, and hence other authors aiming to reuse intermediate results should wait until such snapshot was created. To cope with this limitation, and inspired in Open Source Software development practices, we introduced a \textit{Fork} action\footnote{https://help.github.com/articles/fork-a-repo/} for public, live research objects. Forking a research object means to create a copy of the research object that could be used for testing new ideas without affecting the original research object, or start a new research process based on the forked research object, contributing to speed up research. 

Another fundamental aspect that the original lifecycle lacked was the provisioning of DOIs for research objects. DOIs are an important tool to encourage scientists to change their current way of work to a one based on research objects since they can see the benefits of releasing intermediate results that will be properly credited. DOIs are aligned with the FAIR principles: i) they contribute to the findability of research data and methods, since they are persistent and searchable through a public DOI registry, and ii) they are dereferenceable, meaning that, through a single click, the user will be redirected to a landing page with the main metadata of the research object. Therefore, we extended the lifecycle and associated infrastructure in ROHub\footnote{ROHub is a node of DataCite and an authorized DOI provider.} so that a DOI is automatically generated when a snapshot or and archived research object is released. 

\section{Extracting Content-based Research Object Metadata through NLP}
The reuse of research objects depends to a large extent on their associated metadata. Metadata is key for scientists to evaluate if a given research object produced by someone else is suitable for their own needs, as a whole or partially. Similarly, it is also critical for computer systems, like search engines and recommenders, to automatically collect potentially relevant information through machine-readable annotations.

The research object model supports the generation of metadata enabling research object description from different viewpoints, including lifecycle information (status, evolution, quality checks, authors), resource types (document, workflow, dataset), and information derived from the actual content of such resources, like the specific research areas or the location of the investigation. It can also contain human annotations in titles, labels, descriptions, hypotheses, conclusions and comments. Amongst the different types of metadata, the latter is probably the most descriptive, accurate and valuable in order to obtain a deeper insight on the research since it deals with knowledge directly from the field. However, it formalization requires human involvement and tends to be neglected or embedded in unstructured documents of various formats, like technical reports, presentations or scientific papers. Despite its importance we found that content metadata is scarce for a large number of research objects. From a random sample of 2,500 research objects in ROHub only 800 have such basic content metadata as a descriptive title, with an average character count of 38. In addition, research object descriptions have a typical length of 138 characters, as concise as a Tweet.

\subsection{Semantic Enrichment}
\begin{figure}[tbh]
\centering
  \includegraphics[scale=0.35]{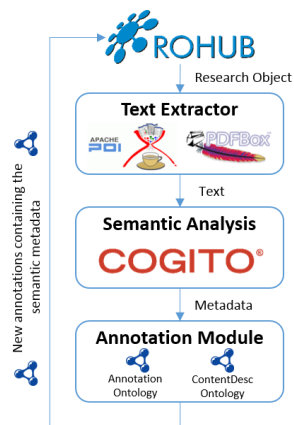}
  \caption{Semantic enrichment process}
  \label{fig:SemEnrichment}
\end{figure}

To alleviate the scarceness of content descriptive annotations and to structure them beyond plain text, we propose to automatically enrich research objects with semantic metadata extracted from human-generated content in the research object, enhancing human and machine readability thus contributing to enable FAIR research and in line with related efforts like the Concept Web Alliance \cite{groth2010anatomy}. The resulting annotations are structured as semantic markup based on a knowledge graph \cite{Uren200614} and included as annotations following the research object model. The enrichment process, depicted in Figure \ref{fig:SemEnrichment}, comprises three main stages: the extraction of text from resources in the research object, the semantic analysis of such text, and the actual generation of semantic metadata.

\subsubsection{Text Extraction}
The enrichment process starts by gathering all the text available within research object resources and human annotations. We process resources in plain text, Microsoft Word and Powerpoint, and Adobe PDF formats, tagged as any of the following types\footnote{Resource type  is assigned upon research object modeling in ROHub.}: Title (\textit{dcterms:Title}), Description (\textit{dcterms:Description}), Document (\textit{wf4ever:Document}), BibliographicResource (\textit{dcterms:BibliographicResource}), Conclusions (\textit{roterms:Conclusions)}, Hypothesis (\textit{roterms:Hypothesis}), ResearchQuestion (\textit{roterms:ResearchQuestion}), and Paper (\textit{roterms:Paper}). We use open source tools to process PDF and Microsoft formats, such as apache PDFBOX and POI. %\footnote{See \url{https://pdfbox.apache.org} and \url{https://poi.apache.org}, respectively}. 

\subsection{Semantic analysis}\label{sec:SemAna}
Research object enrichment builds on the semantic analysis of text\cite{Reeve:2005:SSA:1066677.1067049}, supported by tools such as DBpedia Spotlight\cite{Mendes:2011:DSS:2063518.2063519}, which uses Wikipedia articles as senses to annotate the text, or GATE\cite{Cunningham2011a}, for ontology-based text annotation. Note that this paper focuses on the benefits of semantically annotating research object content beyond the actual tool producing such annotations. So, we will not compare the different alternatives avaliable. In this case we used Expert System's commercial platform Cogito\footnote{\url{http://www.expertsystem.com/cogito}} for convenience but could have chosen a different option. Rather than trying to cover the whole spectrum of metadata specified by the research object model, we focus on a more limited set of annotations supported by Cogito, that describe textual content at the domain level as follows:  

\begin{itemize}
    \item \textbf{Main Concepts} most frequently mentioned in a document. A concept groups words with the same meaning. E.g., \textit{reservoir, artificial lake, man-made lake} are used to refer to \textit{a lake used to store water for community use}.
	\item \textbf{Main Domains}: Fields of knowledge in which the main concepts are commonly used, e.g. \textit{Hidrology} for the words in the former case. 
	\item \textbf{Main Lemmas}: The cannonical form of the most frequent words in the text, e.g., \textit{reservoir, artificial lake, and man-made lake}. A lemma can have different meanings and be associated to more than one concept, e.g. \textit{reservoir} can also refer to \textit{a person, animal, plant or substance in which an infectious agent normally lives and multiplies}.
	\item \textbf{Main Compound Terms}: Most frequent noun phrases\footnote{\url{http://dictionary.cambridge.org/dictionary/english/noun-phrase}}, a group of words in a sentence that together behave as a noun. E.g., \textit{water reservoir} or \textit{hydrochemical element}.
	\item \textbf{Main Named Entities}: Most frequently mentioned named entities, i.e. People, Organizations and Places. E.g., \textit{the black sea} is a place, \textit{UN} is an organization, and \textit{Elizabeth Mary} is a person. 
\end{itemize}

Cogito is built on a knowledge graph (Sensigrafo), where concepts (syncons) are represented as groups of lemmas with the same meaning. Syncons are interconnected through semantic and linguistic relations, like hyperonymy, hyponymy and other properties. The English standard Sensigrafo we used in this work contains 301,582 syncons, 401,028 lemmas and 80+ relation types that yield about 2.8 million links. Among other purposes, Cogito leverages the knowledge contained in Sensigrafo to disambiguate the meaning of a word by recognizing its context.  

\subsubsection{Annotation Generation}   
At the final stage we add the annotations produced by Cogito as research object metadata, following the annotation ontology, which is the standard way to annotate resources in the research object model, and the ContentDesc vocabulary (see \url{https://w3id.org/contentdesc}), which we developed to explicitly link these annotations to the semantics identified by Cogito. We have integrated the semantic enrichment service in ROHub as a nightly daemon, and a collection of semantically enriched research objects is available at \url{http://everest.expertsystemlab.com/browse}, including a search engine built on Solr\footnote{\url{http://lucene.apache.org/solr}}. %The vocabulary provides content negotiation for the most common RDF serializations (rdf+xml, text/turtle, and text/n3) and HTML visualization.

\subsubsection{Semantic Enrichment Example}
The research object \textit{Land Monitoring Change Detecting Step}\footnote{\url{http://sandbox.rohub.org/rodl/ROs/LandMonitoring_Change_Detecting/}} contains a workflow for change detection analysis and includes textual documents describing the hypotheses and conclusions of the analysis. The code excerpt in listing \ref{lst:enrichment} shows the turtle\footnote{https://www.w3.org/TR/turtle} serialization of the semantic annotations added to the research object that were extracted from the textual content.

\lstset{ basicstyle=\scriptsize\linespread{0.7}\ttfamily, columns=fullflexible, xleftmargin=5mm, framexleftmargin=5mm, numbers=left, stepnumber=1, breaklines=true, breakatwhitespace=false, numbers=none, numberstyle=\footnotesize, numbersep=5pt, tabsize=2, frame=lines, captionpos=b,caption={Example of semantic annotations}}

\begin{lstlisting}[float,floatplacement=h,,label={lst:enrichment}]
@base: <.../LandMonitoring_Change_Detecting> .
@prefix skos:  <http://www.w3.org/2004/02/skos/core#> .
@prefix dc:    <http://purl.org/dc/terms/> .
@prefix cdesc: <https://w3id.org/contentdesc/> .

<.../ROs/LandMonitoring_Change_Detecting_Step>
	dc:subject <subject/1302006390>,<subject/280343272>,                 
					       <subject/734754489>,<subject/1557562560>, 
                 <subject/1852089416>,<subject/79018874> .

<subject/1557562560>  a  "cdesc/Concept" ;
        skos:prefLabel  "Segmentation and Reassembly" .

<subject/1852089416>  a  "cdesc/Concept" ;
        skos:prefLabel  "Monitoring" .

<subject/79018874>  a  "cdesc/Domain" ;
        skos:prefLabel  "Geology" .

<subject/280343272>  a  "cdesc/Domain" ;
        skos:prefLabel  "Graphic" .

<subject/734754489>  a  "cdesc/Expression" ;
        skos:prefLabel  "image processing algorithm" .

<subject/1302006390> a  "cdesc/Expression" ;
        skos:prefLabel  "exploitation of image archive" .
\end{lstlisting}

In this example the semantic enrichment added six pieces of metadata stating that the research object content, as defined by the \textit{dc:subject} predicate, mainly refers to concepts (\textit{cdesc/Concept}) "Monitoring" and "Segmentation and Reassembly", which fit in the "Geology" and "Graphic" domains (\textit{cdesc/Domain}). Two of the most frequent compound terms or expressions (\textit{cdesc/Expression}) are “exploitation of the image archive” and “image processing algorithm”. Since the research object actually aims at detecting changes in a region by analysing satellite images and applying different image processing algorithms, the resulting metadata provides a rather accurate summary.

\subsubsection{Assessing the  Relevance of the Semantic Metadata}
We asked members of the organizations participating in our study to answer a questionnaire regarding the new metadata added to the research objects. The objective was to assess the relevance of the annotation types (Domains, Concepts, Named Entities and Compound Terms) with which research objects are enriched against the research object content. In total, 10 researchers participated, who evaluated 19 research objects from their area of expertise and their annotations. %Figure \ref{fig:SurveySemEnrich} summarizes the results. 

% \begin{figure*}[ht]
% \centering
%   \includegraphics[scale=0.7]{surveySemEnrich}
%   \caption{Survey results about the semantic enrichment relevance to the research object content}
%   \label{fig:SurveySemEnrich}
% \end{figure*}

The analysis of the results \cite{gomez2017HumanMachinePartnership} showed that domains and compound terms in general are perceived as relevant to the research object content, while concepts are also relevant but to a lesser extent, and named entities were not found useful by most of the evaluators. Domains are identified by aggregating the domains of all the concepts inferred from the text. Since we are reporting the most frequent domains in the text, erroneously identified domains are left in the long tail of the domain distribution. Compound terms, in turn, explicitly appear as  expressions in the text, hence the high relevance perceived by the participants. 

%Concepts were deemed less useful than expected, with only a slightly positive ratio. However, this is not entirely surprising since word sense disambiguation is still an open problem where state of the art tools such as \cite{Chan07nus-pt:nus-pt} and \cite{tratz2007pnnl} produce f-measure figures \footnote{F-measure is the harmonic mean of precision and recall.} around 0.59 according to SemEval \cite{Pradhan:2007:STE:1621474.1621490}. Plus, we (purposefully) used a standard version of Cogito, without extensions for the Earth Science domain. We also found participants sometimes felt confused when presented the main lemma of the concepts identified by Cogito, which not necessarily was the same as the actual word used in the text (e.g. soil vs. earth). As to the reason why named entities were rated as slightly non relevant, we found that, due to the lack of domain specialization, the system was confused when it came to disambiguate names. For example, the earth observation program Copernicus was confused with the astronomer Copernicus, simply because the former was not known to the system. 

The results showed evidence that automatically produced semantic metadata brings about a positive enrichment of research object descriptions. They also suggest that dedicated user interfaces enabling users to act as curators of the annotations generated may be needed, since a fully automated solution is not feasible yet, given the state of the art in word sense disambiguation. However, we confirmed that a standard, out of the box version of Cogito can produce sufficiently good results for many of the target types of metadata, whose accuracy would be significantly improved, particularly for named entity recognition, with an extended version of Sensigrafo including additional Earth Science knowlege. %Finally, an interesting finding relates to the cognitive gap between how concepts are referred to in the text and semantically equivalent terminological alternatives, and how such gap produces a (negative) effect in the perception of the user. 

\subsection{Research Object Quality}
\label{sec:meta-quality}
Research objects with high quality metadata are more likely to be reused than low quality ones, and in the long term such quality could experience changes, for example when some input file (e.g, an annotation file) becomes unavailable, degrading the overall quality of the research object and introducing decay. Inspired in wet lab practices checklists \cite{hales2006checklist} were proposed as the main tool to assess the quality of research objects through their lifecycle \cite{gomez2013history}. These checklists are made up of statements that specify the required metadata a research object must contain.

A checklist contains the requirements that a research object must fulfill for a given purpose. It is not realistic to have a single set of criteria that fits all situations, i.e. the required metadata when reviewing an experiment differs from that involved in workflow execution. A requirement is a condition about the research object metadata and can be defined as mandatory, desirable, or optional. Requirements are validated through rules that describes how the requirement has to be tested. The most common type of rules are queries over the research object metadata to check for the existence of a particular piece of metadata. 

%For example a requirement about the mandatory inclusion of hypothesis in research objects could be evaluated with a rule that searches for a resource of type Hypothesis in the research object metadata. This search can be implemented by verifying the existence of the triple \lstinline[basicstyle=\footnotesize\ttfamily]{?resource rdf:type roterms:Hypothesis} in the research object manifest, since the semantics of \textit{rdf:type}\footnote{ \url{http://www.w3.org/1999/02/22-rdf-syntax-ns\#}} and \textit{roterms:Hypothesis}\footnote{ \url{http://purl.org/wf4ever/roterms\#}} ensures that \textit{resource} is a Hypothesis. 

%Evaluation and visualization of checklists are supported in ROHub.org\cite{palma2014rohub}, the reference platform for research object management. In the quality tab of the research object view (see figure \ref{fig:wfChecklist}) the user can run and visualize the results of any checklist available in the system. The requirement successfully checked are marked with green ticks, the optional requirements successfully checked are marked with orange ticks and the unsuccessfully ones with orange crosses, and the missing mandatory requirements are marked with red crosses. 

Checklists collect the necessary information to calculate quality metrics about the completeness, stability and reliability of research objects\cite{gomez2013history}. Completeness measures the extent to which a research object satisfies a number of requirements specified in a checklist, stability measures the degree to which the research object completeness remains unchanged, and reliability combines both previous metrics to provide a unique value indicating to what extent the research object is complete and how stable it has been historically. These metrics are visualized in ROHub via an interactive chart displayed after clicking the RO monitoring tool link in the quality tab. %(\hl{NOT SURE IF WE SHOULD INCLUDE THIS FIGURE} see blue chart in figure \ref{fig:wfChecklist}).

%\begin{figure*}
%	\centering
%		\includegraphics[width=0.7\textwidth]{wfChecklist.pdf}
%	\caption{Checklists evaluation and monitoring tool}
%	\label{fig:wfChecklist}
%\end{figure*}
 
%\subsubsection{Checklists for Earth Science}
%We are piloting the adoption of the research object concept in Earth Science with four main communities that embraced the model as a mean to share, reuse, and preserve in the long-term scientific knowledge\cite{EVEREST:D4.1}: 

% \begin{itemize}
% 	\item \textbf{Sea Monitoring}, represented by the Italian Institute of Marine Science (CNR-ISMAR)\footnote{\url{http://www.ismar.cnr.it}}.
% 	\item \textbf{Natural Hazards}, through the UK Natural Hazards Partnership (NHP)\footnote{\url{http://www.naturalhazardspartnership.org.uk}}.
% 	\item \textbf{Land Monitoring}, represented by the European Union Satellite Centre (SatCen)\footnote{\url{https://www.satcen.europa.eu}}.
% 	\item \textbf{Geohazard Supersites and Natural Laboratories} (GSNL)\footnote{\url{http://supersites.earthobservations.org}}, represented by the Italian National Institute of Geophysics and Volcanology.
% \end{itemize}

The hackathons allowed earth scientists to acquire experience with the research object model, create their own research objects and become aware of related benefits for their daily work. Scientists actually proposed specific new types of research objects to encapsulate mainly information regarding scientific workflows, data products, research products, and bibliographic information, which required to design different checklists to assess their quality \cite{EVEREST:D4.3}:  

\begin{itemize}
	\item \textbf{Basic}: This checklist addresses the minimum metadata required for a research object such as title, description, author, and access level. The rest of checklists presented below extend the basic checklist. 
	\item \textbf{Workflow}: This checklist is intended for research objects built with a scientific workflow at the core. It tests metadata such as workflow definition, workflow execution, input and output data (including format and size), and workflow documentation. 
	\item \textbf{Data Product}: This checklist addresses research objects containing mainly data sets. It checks metadata such as the purpose of the data, editor, copyright owner, access level, data format and size. 
	\item \textbf{Research Product}, recommended for research objects dedicated to the analysis of data processing outcomes. It tests metadata such as the purpose, process implementation and input and output data. 
	\item \textbf{Bibliographic}: This checklist is intended for research objects containing mainly bibliographic information such as bibligraphic references or documents that are a relevant to a specific topic. It tests metadata such as the copyright holder, the purpose and access level, and the existence of at least one resource of type Bibliographic resource. 
\end{itemize}

These checklists have been developed and made available in the Earth Science branch\footnote{\url{https://github.com/wf4ever/ro/tree/earth-science/checklists}} in the research object github repository, and can be applied in ROHub to any research object in the Earth Science Domain.

\section{Leveraging Research Object Metadata for Search and Recommendation}
The research object metadata and text extracted from its payload can be leveraged by information retrieval tools that makes them visible to other researchers, thus improving their likelihood to be reused. Mainstream search engines are an important component since they reach a large number of users. ROHub allows web crawlers from Google and Bing indexing the research objects. 

In addition, ROHub provides is own faceted search engine that uses the lifecycle metadata, the user-genetared metadata, and the content metadata generated by the semantic enrichment to ease the browsing of the research object collection. Facets allow the user to filter the collection by selecting specific values in properties (representing the facets) related to the research object (e.g., creator, or creation date). Some of these properties have values linked to a structured knowledge in the form of reference vocabulary (or ontology), such as \textit{research area}, \textit{type of research object}, \textit{state of the life cycle}. Ontologies provides semantics to the property values, and enable semantic inference (e.g., a research object with research area astronomy, is also about space science).

Basic information about research objects is provided to external services through a public search engine interface. It is implemented using the OpenSearch specification \url{http://www.opensearch.org/} which makes it easily adopted by different clients and frameworks. ROHub's OpenSearch interface supports full text search for keyword based scenarios. In order to support finding research objects relevant to specific geographic region a spatial search extension was implemented. It allows usage of spatial intersection queries and returns georss elements \url{http://www.georss.org} in the output document.

Search engines are one of the tools of information retrieval, but not the only ones. Recommender Systems, on the other hand, support exploratory processes and search by example that could help researchers to find research works related to their own. In the following we describe a new recommender system that we developed benefiting of text within research objects and the metadata generated by the semantic enrichment. 

\subsection{Recommender System}
A recommender system\cite {resnick1997recommender} supports exploration when users do not know exactly what to search but have a partial knowledge of e.g. desired characteristics and related examples. Our recommender is content-based\cite {lops2011content}, i.e. user interests are expressed as a collection of research objects and matched against other research objects based on their content. This leverages the research object social dimension through forms of interaction among researchers such as research object coauthoring and citation.

\begin{figure*}[ht]
\centering
  \includegraphics[width=0.6\textwidth]{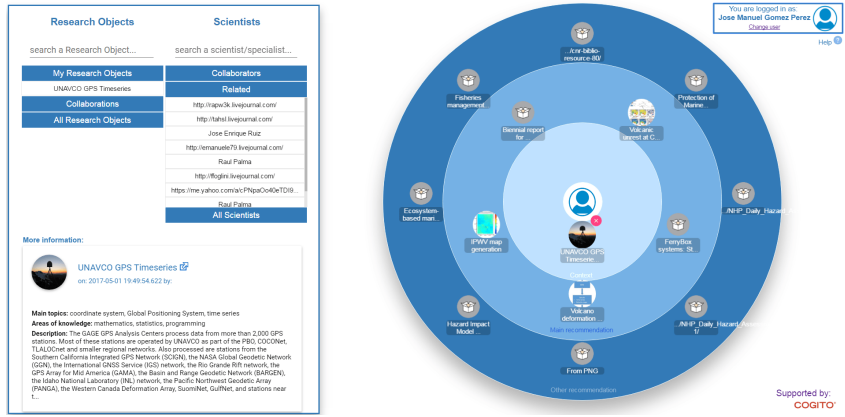}
  \caption{Collaboration Spheres: Recommender system user interface.}
  \label{fig:cs}
\end{figure*}

%\subsection{User Interface}
We implemented a new recommender\footnote{API at \url{http://everest.expertsystemlab.com/home/recommendation-api.html}} based on the results of the experiments reported below, which exploits the metadata generated by the research object semantic enrichment process. The user interface built on top of it is shown in Figure \ref{fig:cs}. The system is accessible from \url{http://everest.expertsystemlab.com/spheres/index.html} and from ROHub (menu \textit{Discover}).   

The user interface follows a visual metaphor designed to facilitate research object sharing and reuse through goal-driven exploration of potentially large collections of research objects. It consists of a navigation panel and information card about the selected research object or scientist on the left-hand side, a set of concentric spheres on the right-hand side, and an authentication box and help option on the upper-right corner. Upon user authentication, the system produces personalized recommendations based on the collection of research objects (s)he authored. Through the navigation panel, the user can search for research objects or community members to be added to the recommendation context. The panel segments the collection of research objects in three subsets in decreasing order of proximity: the research objects authored by the user, those authored by collaborators, i.e. contributors to his or her research objects and the rest. Similarly for community members: collaborators, scientists related topic-wise and others. 

The spheres component serves as a container for both the recommendation context and the recommendation results. Visually, the user is at the center of the spheres. The first sphere around it is an interactive area where the user can drag and drop up to three research objects, scientists (which, processing-wise, act as a proxy to their research objects), or a combination of both from the navigation panel in order to modify the recommendation context. The second and third concentric spheres display the recommendation results. The recommender assigns a score to each resulting research object, indicating its similarity with the recommendation context, which is used to sort the results. The higher the score, the closer to the center.

The usability and user satisfaction of the approach was assessed previously in \cite{DBLP:journals/corr/abs-1710-05604}. Evaluators answered 50 questions\footnote{Questions available at \url{https://sites.google.com/site/spheresquestionnaire/}} aimed at evaluating usability, user satisfaction, perceived usefulness and perceived ease of use. Average usability was 3.95 in a scale of 1 to 5, user satisfaction was 5.61 (1-7), and usefulness and ease of use scored 5.82 in the same scale. 

\subsection{Research Object Similarity}\label{rosim}
Research object recommendation builds on a notion of similarity between research objects in the collection and the ones included in the recommendation context. To calculate this similarity we use the traditional vector space model\cite{Salton:1975:VSM:361219.361220}, whereby documents (i.e. research objects) and interests are mapped to vectors in a multidimensional space where they can be compared using the cosine function as an indicator of similarity between them. Each dimension in this space is weighted according to a predefined weighting scheme\cite{SALTON1988513} and corresponds to a keyword (or other kind of metadata) in the vocabulary that is used in the research object collection. %Can we give some quantifiable description of the vocabulary e.g. in number of terms?

We carried out different experiments to better characterize the similarity measure, with different feature sets used to represent the research objects in the vector space model. The alternatives involved both the keywords extracted from the textual content in the research objects and the semantic metadata generated by the semantic enrichment process. We used the standard TF-IDF\footnote{TF-IDF stands for Term Frequency-Inverse Document Frequency.} as our weighting scheme. Note that the number of research objects in the Earth Science domain is still limited in ROHub since the community is just adopting the paradigm. Therefore we resorted to Wikipedia, where there is a good coverage of articles on Earth Science topics. The belonging of such articles to the domain can be easily determined through the categories assigned to them by the editors. 

\subsection{Experimental Setup}

To generate the evaluation dataset we traversed the Wikipedia category graph starting in the Earth Science category\footnote{\url{https://en.wikipedia.org/wiki/Category:Earth_sciences}}, drilled down three levels in the subcategories, and collected all the articles annotated with these categories. We used DBpedia\footnote{\url{http://dbpedia.org}}, the structured version of Wikipedia, to easily traverse the category graph. In total we harvested 27019 articles that were annotated with 1210 categories. We use such categories as indicators of similarity between articles. For each article we extracted the article title and textual content, discarding all the Wikipedia markup language tags, tables, references, image captions, and infoboxes. %We use  WikiClean\footnote{\url{https://github.com/lintool/wikiclean}} to obtain the plain text for each article. 
Then we created a research object for each article and proceeded to semantically enrich them. %It is worth noting that for this exercise we are using the English standard sensigrafo that was described previously in section \ref{sec:SemAna} without any customization focused on the Earth Science domain. 

To evaluate the similarity measure we use precision at k, a commonly used evaluation metric of ranked results in information retrieval\cite{manning2008introduction}. In our case, precision measures the fraction of research objects identified by the similarity measure that are actually similar to the reference research object. Precision at k is computed on the subset of similar research objects until the k position of the ranked list of similar research objects. We repeated the experiments 10 times and report average precision (p) at 1, 5, 10 and 20.

\subsection{Experiment 1}
In the first experiment we calculated the similarity between a reference research object and the rest in the dataset. From our dataset we selected categories with at least 40 research objects, and randomly selected 10\% of research objects in these categories. In total we assessed the similarity results regarding 2214 research objects under 250 categories. In addition to research objects in the same category, we used a relaxed definition of similarity where we considered as similar research objects also those in neighbor categories, i.e. subsumer (parent), siblings, and children categories. For example, the neighbor categories of \textit{Marine Biology} are the subsumer \textit{Oceanography}, the sibling \textit{Marine Geology}, and the children \textit{Marine Botany}, and \textit{Cetology}. This similarity definition also indicates the variety of related research objects identified by the similarity measure, a desired property in recommender systems.

% Table generated by Excel2LaTeX from sheet 'paper'
\begin{table}[htbp]
  \centering
  \scriptsize
	%\singlespacing
	\tabcolsep=0.11cm
  \caption{Similarity Evaluation for one document}
    \begin{tabular}{lccccc}
    \multicolumn{6}{c}{\textbf{Similarity evaluated on same category}} \\
    \multicolumn{1}{c}{\textbf{Similarty based on}} & \textbf{p@1} & \textbf{p@5} & \textbf{p@10} & \textbf{p@15} & \textbf{p@20} \\
     Concepts and text  & 0,571 & 0,493 & 0,448 & 0,420 & 0,398 \\
     Sem. metadata no NE and text & 0,565 & 0,490 & 0,445 & 0,417 & 0,396 \\
     Sem. metadata and text & 0,569 & 0,490 & 0,445 & 0,417 & 0,396 \\
     Concepts and NE and Text & 0,567 & 0,487 & 0,444 & 0,416 & 0,395 \\
     Text (content+title) & 0,568 & 0,490 & 0,445 & 0,417 & 0,394 \\
     Sem. metadata no NE & 0,480 & 0,415 & 0,378 & 0,355 & 0,339 \\
     Sem. metadata & 0,481 & 0,412 & 0,373 & 0,350 & 0,335 \\
     Concepts & 0,456 & 0,385 & 0,352 & 0,330 & 0,313 \\
     Concepts and NE & 0,456 & 0,384 & 0,347 & 0,324 & 0,307 \\
    \multicolumn{6}{c}{\textbf{Similarity evaluated on neighbor categories}} \\
     Concepts and text  & 0,717 & 0,656 & 0,621 & 0,598 & 0,580 \\
     Sem. metadata no NE and text & 0,718 & 0,654 & 0,620 & 0,597 & 0,579 \\
     Text (content+title) & 0,718 & 0,657 & 0,620 & 0,597 & 0,578 \\
     Concepts and NE and Text & 0,718 & 0,654 & 0,617 & 0,594 & 0,576 \\
     Sem. metadata and text & 0,718 & 0,654 & 0,617 & 0,594 & 0,575 \\
     Sem. metadata no NE & 0,643 & 0,590 & 0,559 & 0,538 & 0,523 \\
     Sem. metadata & 0,639 & 0,578 & 0,548 & 0,527 & 0,513 \\
     Concepts & 0,613 & 0,559 & 0,529 & 0,507 & 0,491 \\
     Concepts and NE & 0,608 & 0,547 & 0,513 & 0,491 & 0,475 \\
    \end{tabular}%
  \label{tab:oneDocSimEval}%
\end{table}%

The experiment results are shown in Table \ref{tab:oneDocSimEval}, with the different approaches sorted in decreasing order by p@20. The best approach in both versions of the experiment was the combination of main concepts (top 10) generated by the semantic enrichment and  textual content of the research object (concepts and text), followed by the combination of all the semantic metadata except named entities and textual content (semantic metadata no NE and text). In general, the combination of semantic metadata plus text seems to produce better results than semantic metadata alone. One interesting observation is that using only semantic metadata the precision values, albeit smaller, are close to other approaches using it in combination with text content. This supports our claim that automatically generated semantic metadata can alleviate the lack of user-generated metadata like research object title or description. Finally, although precision can still be improved, the similarity values evaluated on neighbor categories are promising.     

\subsection{Experiment 2}
While the first experiment addressed one-to-one similarity-based recommendation, the second experiment aims at evaluating the similarity measure when the recommendation context includes the combined attributes of more than one research object. From the dataset, we randomly selected 1000 pairs of research objects where each pair was not annotated under the same category and the path between the categories in the category graph does not include the Earth Science category (since this would make the two resources barely related).

We use the category graph to determine the similarity between research objects by identifying the path connecting the categories of each of the two reference research objects, with the categories in such path as a similarity indicator. For example, if one of the reference research objects falls in the category \textit{Oceanography} and the other one in the category \textit{Marine Botany} we consider as similar research objects those falling in these categories plus the category \textit{Marine Biology} since there exists the path \textit{Oceanography} $\Rightarrow$ \textit{Marine Biology} $\Rightarrow$ \textit{Marine Botany}, where ``$\Rightarrow$'' means hasSubcategory.

We relaxed this definition by considering as similar objects those annotated with a category falling in the subtree whose root is the least common subsummer LCS \cite{wu1994verbs} of the categories associated with the reference research objects. The LCS\footnote{\url{http://www.igi-global.com/dictionary/least-common-subsumer-lcs/41765}} is defined as the most specific common ancestor of two concepts found in a given ontology, and in our case it represents the semantic commonalities of the pair of categories. For example, the LCS of \textit{Marine Biology} and \textit{Ocean Exploration} is \textit {Oceanography}. Similarly to experiment 1 this relaxed definition of similarity is aimed as an indicator of the variety of related research objects that the similarity measure generates. The experiment results are reported in Table \ref{tab:twoDocSimEval}, where the different approaches are sorted in decreasing order by p@20.

Results, in table \ref{tab:twoDocSimEval} show that using text information alone is the best approach when two research objects are used as the basis to obtain similar research objects. Nevertheless, the use of semantic metadata and text does not seem to harm, to a large extent, the precision of the similarity measure. In this experiment we also validated that the use of the semantic metadata without text produces, although smaller, similar results to the ones that we obtain when we have  textual descriptions. The precision values of the similarity metric based on the LCS subtree are a good indicator of the usefulness of the metric in the recommender system when there are more than one research object in the recommendation context.

% Table generated by Excel2LaTeX from sheet 'paper'
\begin{table}[htbp]
  \centering
  \scriptsize
	%\singlespacing
	\tabcolsep=0.11cm
  \caption{Similarity Evaluation for context with two documents}
    \begin{tabular}{lrrrrr}
    \multicolumn{6}{c}{\textbf{Similarity evaluated on categories in the path}} \\
    \multicolumn{1}{c}{\textbf{Similarty based on}} & \multicolumn{1}{c}{\textbf{p@1}} & \multicolumn{1}{c}{\textbf{p@5}} & \multicolumn{1}{c}{\textbf{p@10}} & \multicolumn{1}{c}{\textbf{p@15}} & \multicolumn{1}{c}{\textbf{p@20}} \\
     Text (content+title) & 0,577 & 0,492 & 0,445 & 0,417 & 0,406 \\
     Sem. metadata no NE and text & 0,567 & 0,490 & 0,441 & 0,413 & 0,403 \\
     Concepts and text  & 0,571 & 0,489 & 0,442 & 0,412 & 0,401 \\
     Sem. metadata and text & 0,563 & 0,485 & 0,439 & 0,410 & 0,399 \\
     Concepts and NE and Text & 0,560 & 0,482 & 0,438 & 0,408 & 0,397 \\
     Sem. metadata & 0,458 & 0,388 & 0,347 & 0,321 & 0,309 \\
     Sem. metadata no NE & 0,448 & 0,387 & 0,343 & 0,321 & 0,308 \\
     Concepts & 0,411 & 0,355 & 0,321 & 0,299 & 0,287 \\
     Concepts and NE & 0,416 & 0,353 & 0,313 & 0,291 & 0,281 \\
    \multicolumn{6}{c}{\textbf{Similarity evaluated on categories in LCS subtree }} \\
     Text (content+title) & 0,740 & 0,677 & 0,643 & 0,626 & 0,618 \\
     Sem. metadata no NE and text & 0,732 & 0,677 & 0,641 & 0,623 & 0,616 \\
     Concepts and text  & 0,736 & 0,678 & 0,641 & 0,621 & 0,613 \\
     Sem. metadata and text & 0,725 & 0,674 & 0,637 & 0,618 & 0,610 \\
     Concepts and NE and Text & 0,724 & 0,673 & 0,636 & 0,615 & 0,607 \\
     Sem. metadata no NE & 0,657 & 0,605 & 0,573 & 0,555 & 0,543 \\
     Sem. metadata & 0,655 & 0,600 & 0,571 & 0,546 & 0,539 \\
     Concepts & 0,617 & 0,583 & 0,549 & 0,530 & 0,520 \\
     Concepts and NE & 0,614 & 0,576 & 0,535 & 0,515 & 0,506 \\
    \end{tabular}%
  \label{tab:twoDocSimEval}%
\end{table}%

\section{Earth science interfaces for research objects}
\label{infra}
Enhancing traditional research practices with FAIR-enabled capabilities based on research objects requires specialized user interfaces that integrate the governance capabilities provided by research objects with existing tools already used by earth scientist in their daily work. In doing so, we need to keep a delicate balance, pushing the boundaries of what is now possible with the current tools (i.e. adding new functionalities) while maintaining the familiarity with current interfaces and user experience. 

In this section, we illustrate how this challenge has been addressed for different communities of scientists with specific needs and goals. The user interfaces and applications selected to that purpose include: the ROHub portal, the main front end for domain-independent research object lifecycle management sitting on top of the RO API; a Virtual Research Environment for vertical communities of scientists, in disciplines like sea monitoring and volcanology; and domain-specific applications dealing with time series data in the ecology and biodiversity domain.

\subsection{ROHub Portal}
The ROHub portal is the generic front-end for research object management that provides an advanced, life cycle management-oriented, tool exposing the full set of research object management capabilities to scientists. It is intended for users who are already familiar with research objects, or who would like to analyze and manage research objects in at a finer grain of detail. Hence, it provides great flexibility and access to all possible operations at a granular level. In contrast, Virtual Research Community (VRC) portals for example (see section \ref{vrc}), provide scientists with access to composite custom-built operations at a higher level of abstraction. So, while in ROHub portal, the user may need to perform multiple individual operations to build a research object (create, annotate, add resources, etc.), the VRC portals encapsulate all these operations in a single, custom-built process.

\begin{comment}
\hl{Raul, we need a feature-based description rather than a technical one.}
\st{The ROHub  portal is based on Play framework providing a lightweight, stateless, web-friendly architecture, and AngularJS, a structural framework for dynamic web apps. The combined usage of these technologies enables creating a modular web application with a set of visual components that can be easily reused in other applications.}
\end{comment}
The portal integrates and provides access to different research object services, including the core services provided by ROHub back-end for their creation, storage, access and maintenance, the management of their lifecycle, and their preservation, as well as added-value services like notification, transformation of workflows into research objects, quality and stability assessment, metadata enrichment, rating and exploratory search.

\begin{figure}[t]
\centering
  \includegraphics[width=0.48\textwidth]{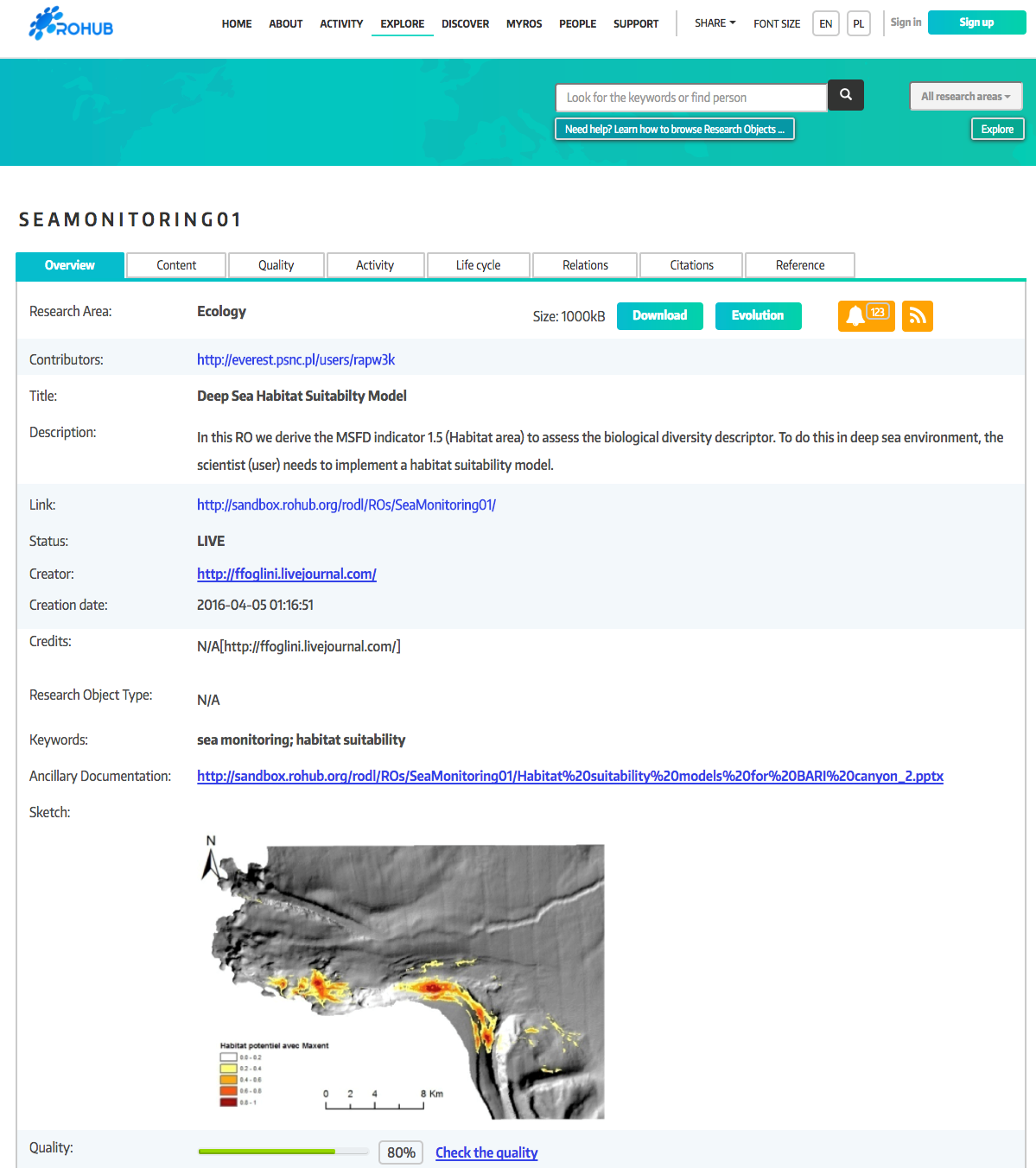}
  \caption{ROHub Portal}
  \label{fig:rohub-portal}
\end{figure}

\subsection{Community-Oriented Virtual Research Portals}\label{vrc}
Earth Science needs to address a variety of challenges. Among them, climate change is probably the most known topic because of its direct link to the increase of the average global temperature, but many others exist, including marine litter, air pollution, flooding and volcanic eruptions. This implies an increasing demand of data and information management capabilities to provide evidence, understand causes and monitor effects. The EVER-EST\footnote{http://vre.ever-est.eu} virtual research environment (VRE) provides the different communities of earth scientists with virtual research community (VRC) portals offering custom services and tools targeted to ease work in community specific tasks. To support collaborative research across institutional and discipline boundaries, the VRE and VRC online portals use the innovative concept of research objects to draw together research data, models, analysis tools and workflows as well as to manage and preserve the full research cycle. These interfaces abstract the research object vocabulary and details from the user, providing custom-built access to the core research object management capabilities in a simple and transparent manner. Currently there are four VRC portals - Land Monitoring\footnote{http://vre.ever-est.eu/landmonitoring/}, Natural Hazards\footnote{http://vre.ever-est.eu/naturalhazards/}, Sea Monitoring\footnote{http://vre.ever-est.eu/naturalhazards/} and GeoHazards Supersites\footnote{http://vre.ever-est.eu/supersites/} - which can be accessed from the VRE, each pre-configured with the associated domain-specific data and services. 

\begin{figure}[t]
\centering
  \includegraphics[width=0.48\textwidth]{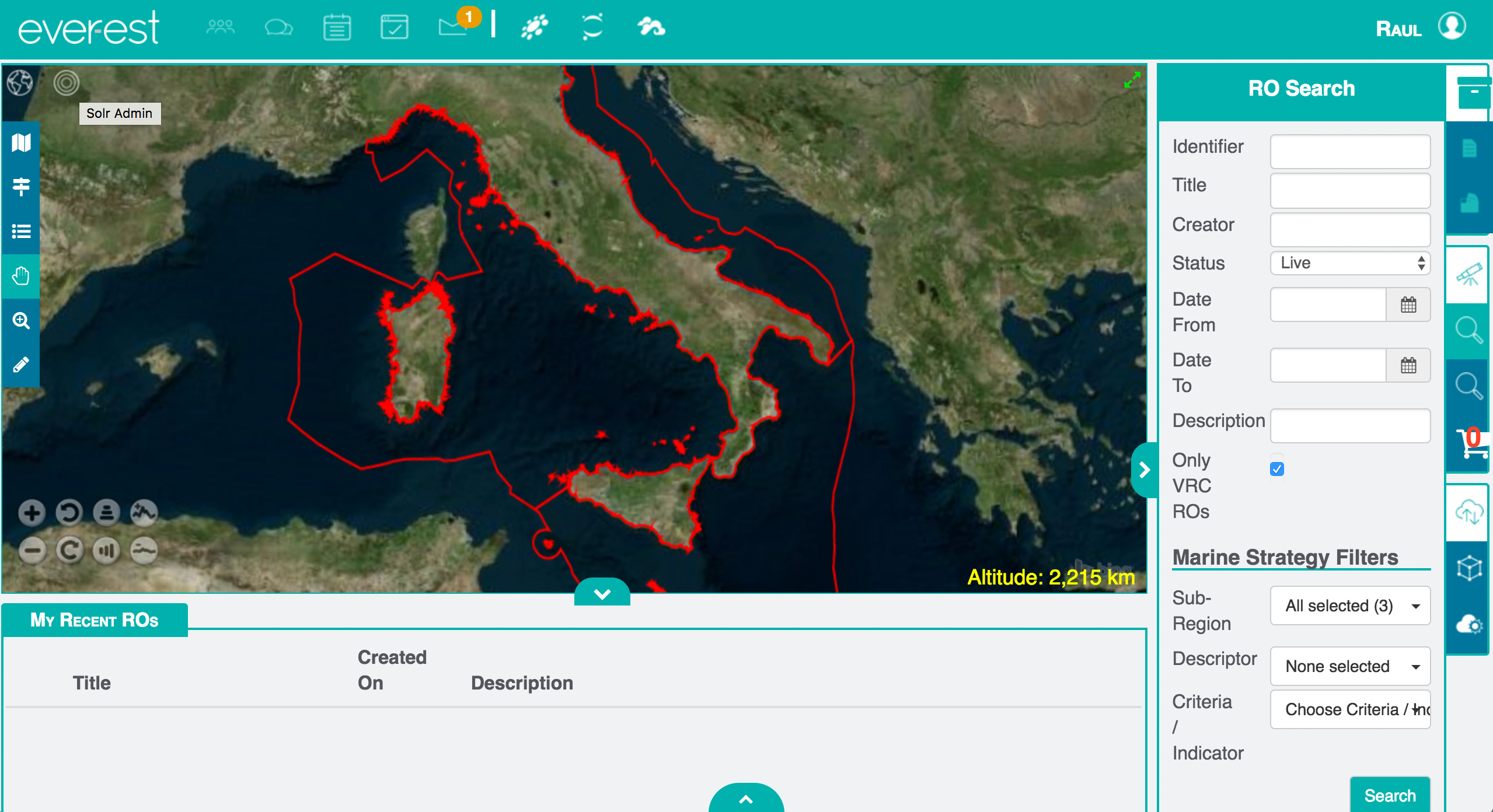}
  \caption{Sea Monitoring VRC Portal}
  \label{fig:vrcportal}
\end{figure}

The VRC portals design reflects the User Interface (UI) and User eXperience (UX) needs shared among the Earth science communities: the 3D virtual globe, the most natural playground for an Earth Scientist to perform his/her activity, plays the central role and provides interactive tools to manage the full research cycle and enable direct interaction and visualization with research data. The toolbar on the right hand side (see figure \ref{fig:vrcportal}) is the research pad that collects and enables features related to research objects and other tools that are commonly used by Earth scientists:

\begin{itemize}
\item \textbf{Research object services:} include basic research object functionalities (e.g. create, edit, annotate, etc), research object lifecycle management, metadata management or resource management. 
\item \textbf{Data discovery:} provides a search box to define search criteria both for Earth Observation datasets (e.g. Sentinel data, Datacube, Co.CO.NET., etc) and Research Objects based on OGC Open Search standard interface. 
\item \textbf{Cloud services:} enable access to the private storage area (i.e. Seafile) and to three macro categories of processing services, namely
\begin{itemize}[leftmargin=.15in]
\item \textbf{Workflow services,} to discover and execute scientific workflows by a generic workflow manager (e.g. Taverna server)
\item \textbf{Virtual Machines,} to provide %Earth scientists with 
access to existing cloud resources, i.e. virtual machines, while enabling VRC administrators to manage them.
\item \textbf{Web Processing Services (WPS),} to facilitate the integration and execution of existing geospatial processes available as web services. 
\end{itemize}	
\end{itemize}

\subsection{Time Series Data Analysis in Ecology and Biodiversity}

Nowadays measuring the causes and effects of environmental change and how ecosystems are affected is a main concern for society and researchers. Scientists working on this problem often need to deal with data from different providers each of one serving the data they are specialized on. Scientists need to compare slices of time series data of different sensors and systems, keeping track of the provenance information that enable others to reproduce the experiments and reuse the results. 

\begin{figure}[tbh]
\centering
  \includegraphics[width=0.48\textwidth]{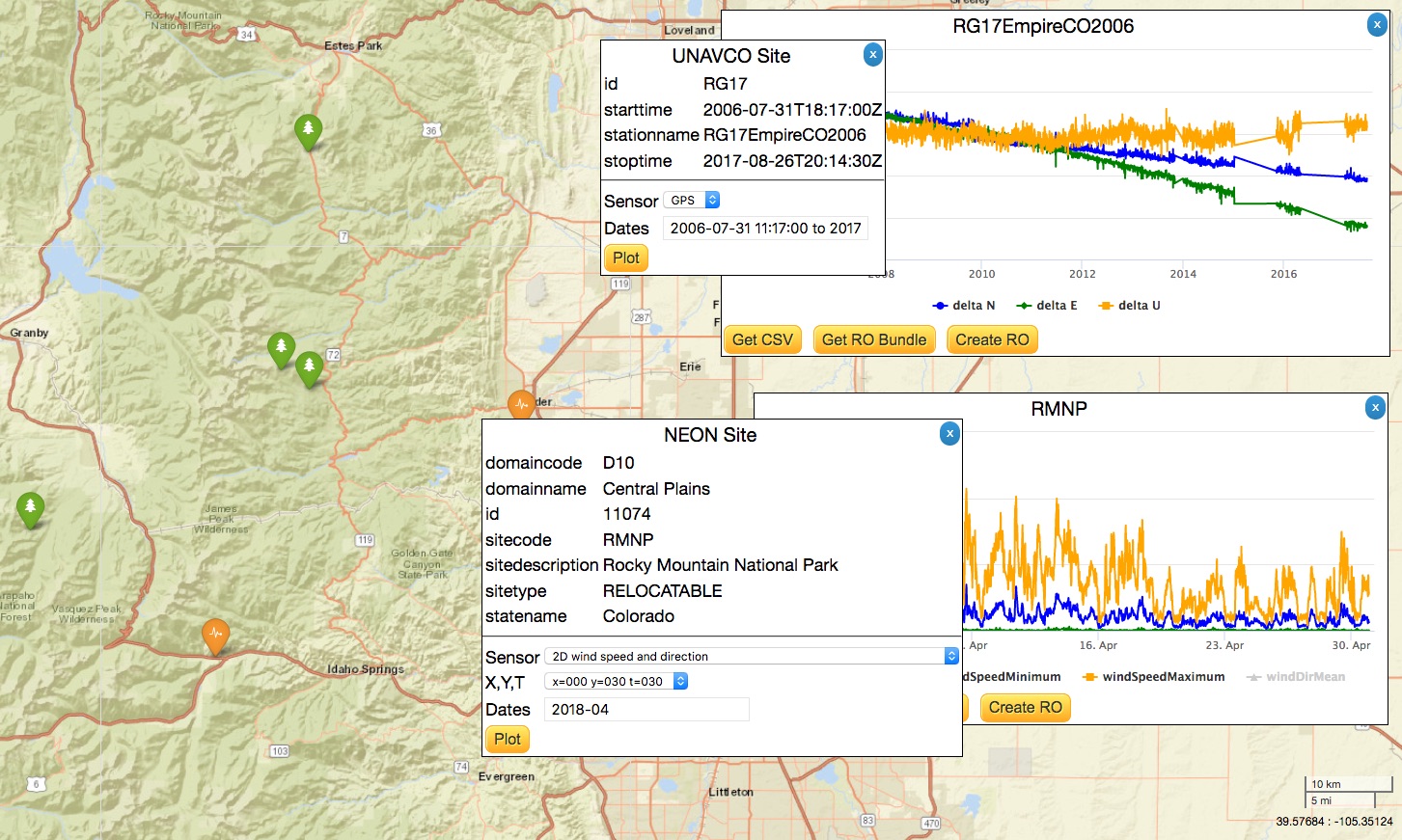}
  \caption{Web application to view UNAVCO and NEON time series.}
  \label{fig:SaviWeb}
\end{figure}

To support scientist interested in ecological processes we have developed and interactive web-based prototype application \footnote{\url{https://firemap.sdsc.edu/savi/map.html}} (see Figure \ref{fig:SaviWeb})  that integrates time series from UNAVCO\footnote{\url{https://www.unavco.org/}} and National Ecological Observatory Network (NEON)\footnote{\url{https://www.neonscience.org/}} sensors, and produces workflow-centric research objects. The UNAVCO stations record GPS positions while sensors in NEON towers provide multiple types of data, e.g., wind speed, humidity, etc., at different time resolutions. Users can plot and download time series data by selecting the station, sensor type, and time range.

Time series from UNAVCO and NEON are accessible from REST services. Since UNAVCO and NEON provide data in different formats, a workflow was developed in the Kepler Scientific Workflow System \cite{Altintas2004kepler} to perform the REST queries and convert the results into GeoCSV \cite{GeoCSV}.  A Kepler workflow consists of executable components, called ``actors'', linked together based on data dependencies to form an overall application. The workflow for this application includes the actor to perform REST queries, and the R actor to convert data into GeoCSV.

After selecting time series from one or more sensors, a research object may be created to encapsulate the data and process used to create it. The research object includes a GeoCSV file containing the time series along with the instance of the Kepler workflow, which contains the parameters used to create the GeoCSV such as sensor location and time range. This workflow may be re-executed
to produce the same time series data. The research object may either downloaded or shared on ROHub.

\section{Community Adoption}
We are still in early stages of the process to build a FAIR community of earth scientists that leverage research objects for the management, sharing and publication of their research and/or operational work on a normal basis. Nonetheless, the infrastructure is solid and we count with a considerable international community of early adopters, fundamentally distributed over Europe and the USA but also with some participation from Australia. 

%As shown above, the different members of such community have already built a good number of custom applications for FAIR research on top of the ROHub infrastructure and are moving their work practices to those inspired by research objects and the FAIR principles. 

The different user interfaces built on top of the ROHub infrastructure presented in Section \ref{infra} have encouraged community members to move their work practices to those inspired by research objects and the FAIR principles. As a matter of fact, our early adopters are already producing and exploiting high quality research objects in both manual and automatic ways. As described in previous sections, research objects are indeed enabling these communities to adopt the FAIR principles in their scientific work: they are modeled based on interoperable ontologies, described with rich and expressive metadata, citable in scholarly communications, visible and discoverable from the Web and via recommendation systems, and ultimately, reusable (see Table \ref{tab:faircomponents}). 

Yet, to better understand the use of the infrastructure by our community of scientists, to obtain a deeper insight and to facilitate the sustainability and continued growth of the community, we have implemented a number of mechanisms to monitor and measure performance. In this section, we provide an account of current progress stemming from quantitative data and related indicators.

\subsection{Featured Research Objects}\label{ro_ex}
Our early adopters increasingly use research objects and the associated infrastructure as part of their daily activities. After gaining a good understanding of the research object paradigm and the supporting technologies, key members of the community created a set of representative research objects for their area. We refer to the resulting research objects as Golden Exemplar Research Objects (GERO)\footnote{\url{http://everest.expertsystemlab.com/\#Golden Exemplars}}. These are particularly curated and representative research objects that allow  demonstrating the feasibility and utility of research objects to manage and share data, models and results of the daily work in Earth Science. Next, we select some of these golden exemplars from two of these communities, to further illustrate this approach:

\bigbreak
\noindent
\textbf{\textit{Sea Monitoring}}
	\begin{itemize}
        \item \textbf{Detection of trends in the evolution of invasive jellyfish distribution}, a workflow-centric research object that produces explicit geographical information concerning the evolution and distribution of alien species based on Jellyfish sightings.
        \item \textbf{Digitalization of historical Venice lagoon maps}, a data-centric research object with information on natural environmental and anthropogenic changes.
		\item \textbf{Deep Sea Habitat Suitability Model}, a workflow-centric research object to derive the Marine Strategy Framework Directive MSFD indicator 1.5 to assess the biodiversity descriptor.
	\end{itemize}
% 	\item Natural Hazards
% 	\begin{itemize}
% 		\item \textbf{Hazard Impact Model}, for surface water flooding simulation and early warning systems.
% 	\end{itemize}
% 	\item Land Monitoring
% 	\begin{itemize}
% 		\item \textbf{Land Change Detection}, which aims at detecting anomalies in time series of satellite images acquired on specific locations and their correlation with social sensing sources and other geotagged information.
% 	\end{itemize}
\textbf{\textit{Geoscience Research}}
	\begin{itemize}
        \item \textbf{IPWV on Iceland}, a workflow-centric research object that automatizes the generation of a map of the precipitable water content on Iceland by using MODIS satellite data.	
        \item \textbf{March 2018 reports at Mt Etna}, a bibliographic research object containing all reports from March 2018 describing the weekly volcanic activity of Mt Etna from the multi-parametric monitoring stations.
		\item \textbf{Volcano Source Modeling (VSM)}, a workflow-centric research object containing the VSM methods and related resources used to obtain results of the geodetic inversion of the 2011-2013 InSAR data at Campi Flegrei (Italy) due to the action of a deep magmatic source.
         \item \textbf{UNAVCO GPS Position Timeseries}, a workflow-centric research object encapsulating a kepler workflow that calls a GPS position timeseries webservice provided by UNAVCO, processes the stream of data, and plots the north, east, and vertical offsets relative to a reference position. 
     \end{itemize}

In addition to these manually crafted, high-quality research objects, we also generated through an automatic process over 500  bibliographic research objects\footnote{\url{http://everest.expertsystemlab.com/\#Generated}} (AGROs - Automatically Generated Research Objects) exposing gray literature periodically released by these institutions, and bibliographic references of interest for the community.

\subsection{Key Performance Indicators}
% * <jose.manuel.gp@gmail.com> 2018-05-25T10:19:15.283Z:
% 
% > \subsection{Key Performance Indicators}
% This whole section should be better linked to table 2, where a mapping between the approach and FAIR principles is provided.
% 
% ^.
We have defined a set of key performance indicators (KPIs), consisting of measurable values, that allow us to: i) assess the success regarding the community adoption of research objects and related technologies; ii) estimate the extent to which this work is contributing to improve the currently limited compliance with the FAIR principles in Earth Science communities; and iii) to identify and analyze usage trends. For each of these KPIs, we defined a target for the six-month period Apr-Sep 2018. The targets were defined with the feedback of key community members regarding their experiences and expectations about research objects and their daily work. Thus, starting from April 2018, KPIs are measured monthly and compared against the targets to assess the progress and to draw conclusions. 

%and started collecting the measured values monthly since the 2nd quarter of this year. The targets were defined with the feedback of key community members regarding their experiences and expectations about research objects and their daily work. Each month we compare the measured values with the targets to assess the progress and to draw conclusions.

The KPIs are measurable via the ROHub platform, which integrates multiple added-value services and serves different client applications (see Section \ref{infra}). Table (\ref{table:kpis}) presents the KPIs, with the target values for the six-month period Apr-Sep 2018, and the last measured values (May 2018). 

As we can observe from the table, we have already reached a few targets, including number of GEROs, number of AGROs and percentage of research object views. Reaching the targets in the number of golden and automatically generated research objects is a good indicator related to community adoption, even though the overall number (GEROs+AGROs+others) is still slightly below the target. More importantly, having already such significant number of research objects is an improvement in the FAIR level of these communities. Concretely, now over 3500 data and other research artifacts are FAIR enabled via almost 750 research objects (see discussion in Section \ref{sec:fair-research}). In fact, reaching the target in the percentage of views can be considered as an evidence indicating that resources are findable and accessible (first two rows in Table \ref{tab:faircomponents}).  
% * <jose.manuel.gp@gmail.com> 2018-05-25T10:14:44.782Z:
% 
% > many data artifacts
% Quantify.
% 
% ^.
% * <jose.manuel.gp@gmail.com> 2018-05-25T10:11:22.038Z:
% 
% how were such target defined and fixed? The way it is now, it looks a little arbitrary.
% RAUL: Updated a little, but in truth most targets were fixed arbitrarily if you remember
% 
% ^.

Table \ref{table:kpis} also shows that some KPIs are still below the target. However, in most cases the values measured are not so far from the targets and a steady increase has been observed; thus, we are confident that the targets will be reached by the end of September 2018. 
%we believe there is still enough time to reach the targets %for the six-month period (end of 3rd quarter). 

%In most cases this is not worrisome as we still have more than a quarter ahead (67\% of the six-month period) and the values are not so far from the targets. 
% * <jose.manuel.gp@gmail.com> 2018-06-16T21:31:00.572Z:
% 
% > enough time to reach the targets for the first period (end of 3rd quarter)
% Enough time before what? What is that deadline about??  clarify or remove the comment.
% 
% ^.
%RAP: I changed to use period apr-sep 2018 everywhere to make it more clear
For example, having still four months to go (i.e., 66 \% of the period Apr-Sep 2018), the number of resources managed by Earth Science communities through research objects is already at 36 \% of the target, %64\% below the target, 
while the average quality of research objects is only between 2 and 17\% below the target. Note that quality-related measurements take into account conditions like whether or not the data and associated research are well described (with rich, machine-readable metadata) or that resources are accessible, all of them key factors in terms of compliance with the FAIR principles (first three rows in Table \ref{tab:faircomponents}). The fact that quality measures are almost aligned with the target values is a good indicator, showing evidence of convergence towards FAIR among the communities. 
% * <jose.manuel.gp@gmail.com> 2018-06-16T21:32:58.006Z:
% 
% > 64\% below the target
% This is HUGE! Rephrase
% 
% ^.RAP DONE

Nonetheless, indicators of reuse (research objects downloads and forks) are still far from the target and we have increased our efforts in analyzing how to raise such values. For instance, a better understanding is needed about how to encourage earth scientists to increase sharing by reusing or repurposing existing results rather than by carrying out their research from scratch. Limited reuse values also indicates the need to provide earth scientists with means to simplify such tasks, lowering the technical entry barrier. Tooling support to enable proper credit to previous work, i.e. through persistent identifiers and enforcing automatic citation, is also key in this regard (last row in Table \ref{tab:faircomponents}). Although such mechanisms are already available in ROHub (e.g. release of research objects with DOIs, research object fork and automatic citation to the source), our analysis seems to indicate some lack of awareness about such functionalities among user scientists. 

Furthermore, we have recently implemented in ROHub mechanisms that on the one hand enable scientists to express a subjective notion of quality about particular research objects and on the other hand keep account of the social impact of a research object among the user communities. Although the amount of data available to this purpose is still limited, we observe a trend indicating a correlation between research object reuse and their popularity. Frequently reused research objects have better ratings and reviews, and are favorited more frequently. As part of our awareness work, such features are now making their way into the user communities. Follow up work in this direction includes mechanisms to highlight or rank scientists depending on the reputation they earned based on the impact (rates, likes, views), reuse (downloads, forks) and quality of their research objects.
% * <jose.manuel.gp@gmail.com> 2018-05-25T10:16:52.802Z:
% 
% > In most cases this is not worrisome as we have still more than a quarter ahead (67\% of the six-month period) and the values are not so far from the targets
% This sounds like you are speaking in terms of project schedule, which is  not correct and raises questions about sustainability. Update. Also, avoid the use of terms like "worrisome".
% 
% ^.
% RAUL: Updated

\begin{table}[!h]
\centering
\caption{Key performance indicators: targets (September 2018) against measures (May 2018)}
\label{table:kpis}
\resizebox{\columnwidth}{!}{%
%\begin{tabular}{ | p{0.5\columnwidth} |
%                   p{0.1\columnwidth} |
%                   p{0.1\columnwidth} |
%                   p{0.1\columnwidth} | 
%                   p{0.1\columnwidth} |
%                   p{0.1\columnwidth} |} 
%\begin{tabular}{llllll}
%\begin{tabular}{ p{8cm}  p{3cm} | l | l | l | l |}
\begin{tabular}{|l|l|l|l|l|l|}
\hline
\multicolumn{2}{|c|}{\textbf{Key Performance Indicator (KPI)}}                                                & \multicolumn{1}{c|}{\textbf{Target}} & \multicolumn{1}{c|}{} & \multicolumn{2}{c|}{\textbf{Measured}} \\ \hline
\multicolumn{2}{|l|}{\multirow{3}{*}{\begin{tabular}[c]{@{}l@{}}Number of research objects \\ implemented in Earth Science\end{tabular}}}                    & GEROs                                & 8                     & GEROs               & 16               \\ \cline{3-6} 
\multicolumn{2}{|l|}{}                                                                                        & AGROs                                & 500                   & AGROs               & 512              \\ \cline{3-6} 
\multicolumn{2}{|l|}{}                                                                                        & Overall                                & 1000                  & Overall               & 748              \\ \hline
\multicolumn{2}{|l|}{\begin{tabular}[c]{@{}l@{}}Number of Earth Science resources\\ managed by the communities\end{tabular}}                      & Total                                & 10000                 & Total               & 3563             \\ \hline
\multicolumn{2}{|l|}{\multirow{3}{*}{\begin{tabular}[c]{@{}l@{}}Average quality of \\ Earth Science research objects\end{tabular}}} & GEROs                                & 95\%                  & GEROs               & 93\%             \\ \cline{3-6} 
\multicolumn{2}{|l|}{}                                                                                        & AGROs                                & 90\%                  & AGROs               & 73\%             \\ \cline{3-6} 
\multicolumn{2}{|l|}{}                                                                                        & Released                             & 85\%                  & Released            & 72\%             \\ \hline
\multirow{5}{*}{\begin{tabular}[c]{@{}l@{}}Impact of Earth Science \\ research objects\end{tabular}}             & \multirow{2}{*}{Views}                & GEROs                                & 100\%                 & GEROs               & 100\%            \\ \cline{3-6} 
                                                                      &                                       & AGROs                                & 40\%                  & AGROs               & 99\%             \\ \cline{2-6} 
                                                                      & \multirow{2}{*}{Downloads}            & GEROs                                & 80\%                  & GEROs               & 44\%             \\ \cline{3-6} 
                                                                      &                                       & AGROs                                & 25\%                  & AGROs               & 2\%              \\ \cline{2-6} 
                                                                      & Forks                                 & Total                                & 25\%                  & Total               & 1\%              \\ \hline
\end{tabular}
}
\end{table}

\subsection{Web analytics}
\begin{figure}[!h]
\centering
  \includegraphics[width=0.48\textwidth]{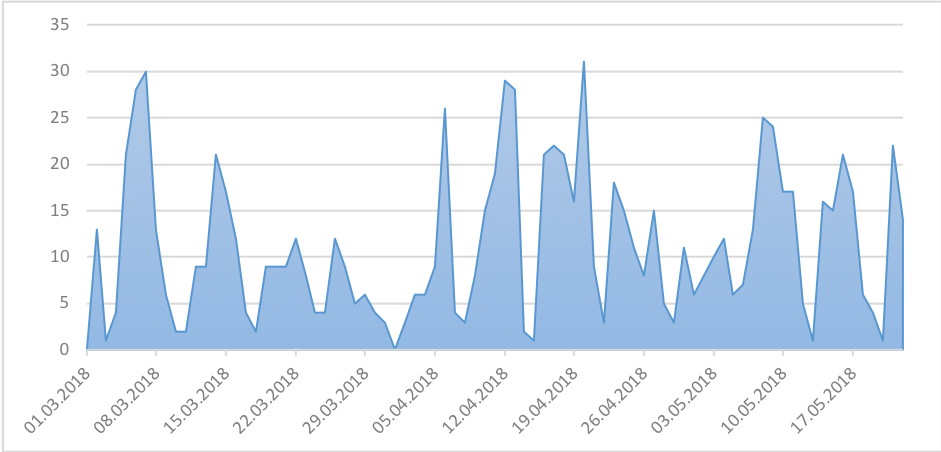}
  \caption{ROHub web traffic: users per day since March 2018}
  \label{fig:rohub-users}
\end{figure}

Another mechanism that was put in place to monitor and to get insights about the adoption of research objects and related technologies is the tracking and reporting of ROHub web traffic using Google Analytics. We started tracking the ROHub Web site since March 1st 2018, and have already collected enough information to discover some patterns. For instance, figure \ref {fig:rohub-users} depicts the number of users visiting ROHub per day, where we can observe multiple peaks. After analyzing these peaks, we see that many of them coincide with the dates of dissemination or demonstration events, which indicates interest from the target communities, e.g., GeoVol (latin american workshop on volcanology) 7th-9th March, or EGU (European Geosciences Union) 9th-12th April. It is worth noting that since the beginning of the track history (83 days including weekends) only one day did ROHub not get any visit: Sunday 1st April (Easter). 

\begin{figure}[!h]
\centering
  \includegraphics[width=0.48\textwidth]{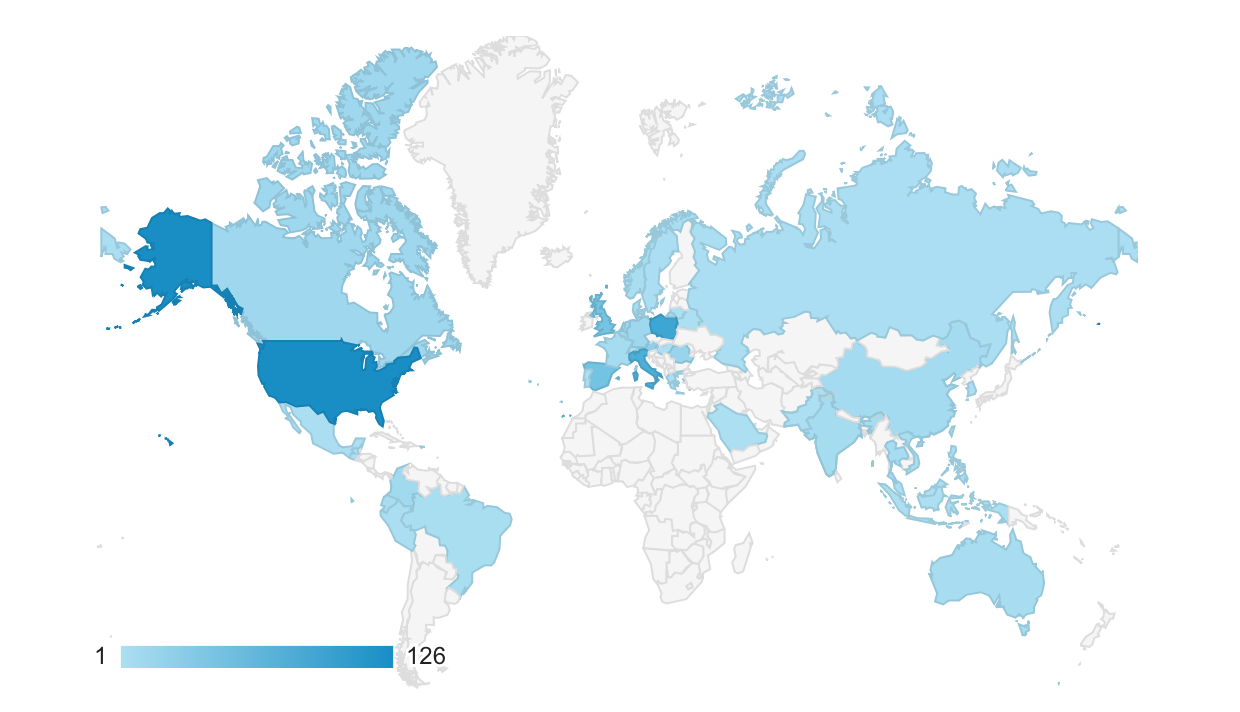}
  \caption{ROHub web traffic: users per country since March 2018}
  \label{fig:rohub-location}
\end{figure}

Regarding the number of users per country, the USA is in first position, with about 23\% of the share (see Figure \ref{fig:rohub-location}). Although we have engaged some Earth Science communities there, this was an interesting finding. The second country is Poland (where ROHub is developed), followed by Italy (where two other important Earth Science communities are located), Spain (where another Earth Science community and a key technical partner are located), and the UK (where another Earth Science community is located).

Though possibly anecdotic, it is interesting to point out  %are the time of the day, and the day of the week with more traffic. We found out
 that the busiest time of day is usually around noon, being 14:00 the busiest hour (based on the number of sessions), followed by 12:00, 11:00 and 15:00. This indicates that the busiest hour is right after lunch in Europe (CET time) and early morning in the United States (Eastern time), which seems to indicate that scientists actually access the platform as part of their daily routine. Regarding the busiest day of the week, we found no significant difference between working days, also indicating usage of the platform as part of the daily work activities.

\section{Conclusions}
In this paper we described the journey we went through to build a FAIR research environment for Earth Science around research objects. Transforming a data-intensive scientific community like this to use FAIR principles is a continuous and multidisciplinary effort that must be supported by methods, models and tools, while engaging early adopters from these communities to produce a critical mass of FAIR content that encourage their peers to adopt this new paradigm of work, leading to the establishment of a virtuous circle of FAIR data sharing and reuse. 

Our work aimed at building upon the research object model a set of tools that ease the generation of research objects while increasing their likelihood to be reused by other researchers. Therefore our focus was on vocabulary extensions, automatic generation of metadata and quality assessment, search engines and recommender systems, digital object identifiers, and tailored user interfaces that incorporate earth science datasets, time-series data management and geolocalization. The key performance indicators to monitor the health of the research community of earth scientist working with research objects are in place. The challenge for the future is to enlarge the user community and leverage the experience gained with earth scientists to encourage other research communities to make the transition to a FAIR data interchange.

\section*{Acknowledgements}
We gratefully acknowledge EU Horizon 2020 for research infrastructures under grant EVER-EST-674907.

\section*{References}

\bibliography{biblio}

\end{document}